\documentclass{article}

\usepackage[main,final]{neurips_2026}

\usepackage[utf8]{inputenc} 
\usepackage[T1]{fontenc}    
\usepackage{hyperref}       
\usepackage{url}            
\usepackage{booktabs}       
\usepackage{placeins}
\usepackage{amsfonts}       
\usepackage{nicefrac}       
\usepackage{microtype}      
\usepackage{xcolor}         
\usepackage{natbib}
\usepackage{booktabs}
\usepackage{graphicx}
\usepackage{caption}
\usepackage{multirow}
\usepackage{amsmath}
\usepackage{amssymb}
\usepackage{algorithm}
\usepackage{algpseudocode}
\usepackage{todonotes}
\usepackage{bbm}
\usepackage{wrapfig}
\usepackage[table]{xcolor}
\definecolor{gooddelta}{RGB}{0,130,0}
\definecolor{baddelta}{RGB}{180,0,0}
\newcommand{\gd}[1]{\makebox[0pt][l]{\hspace{0.3em}\textcolor{gooddelta}{\scriptsize(#1)}}}
\newcommand{\bd}[1]{\makebox[0pt][l]{\hspace{0.3em}\textcolor{baddelta}{\scriptsize(#1)}}}
\newif\ifcomments
\commentstrue
\ifcomments
     \providecommand{\mert}[1]{{\color{olive}{[mert: #1]}}}
    
\else
    \providecommand{\mert}[1]{}

\fi

\title{Spectral Feedback for Test-Time Alignment of Protein Diffusion Models}

\author{
Shai Dickman \qquad
Mert Cemri \qquad
Landon Butler \qquad
Kannan Ramchandran \\
Department of Electrical Engineering and Computer Sciences\\
University of California, Berkeley\\
}

\begin{document}

\maketitle

\begin{abstract}

Discrete diffusion models have become leading architectures for inverse protein folding, which involves complex non-causal interactions. Test-time alignment is a useful technique in protein design to guide these models at inference to generate proteins with desirable attributes. Alignment methods for discrete diffusion models have primarily focused on steering the diffusion reverse process, either by influencing token logits or by selecting favorable sequences at intermediate steps to eventually yield high-reward samples. These approaches largely treat inference as a unidirectional process, lacking effective mechanisms for revisiting undesirable token selections. We introduce \textbf{Spectral Feedback}, an algorithm that selects edit-positions in a feedback loop, allowing the model to iteratively correct its own generations. This approach leverages the masking structure of discrete diffusion models by re-masking and re-sampling tokens, analogous to image editing methods that reintroduce noisy latents and re-run the reverse diffusion process. While prior alignment methods focus on \textit{what token labels} to assign to maximize a target reward, we instead treat \textit{which tokens} to revisit as the central alignment problem. Selecting edit-positions is challenging because edit effects are interdependent: the impact of modifying one token depends on which others are edited simultaneously. We define an edit-set as a set of positions to edit by re-masking and re-sampling the corresponding tokens in a sequence. Motivated by prior work on sparse interactions in biological systems, we find empirically that edit-set value functions for protein inverse folding admit sparse Fourier representations. This structure enables Spectral Feedback to efficiently learn and optimize the value functions for edit-position selection. Spectral Feedback is model-agnostic and can be applied to pre-trained, test-time aligned, and fine-tuned diffusion models without modifying the underlying generative processes. Applied to inverse folding with a protein stability reward oracle, Spectral Feedback increases the proportion of stable proteins by 32.3 percentage points over a pre-trained model, 24.8 over Best-of-10, and 5.8 over a state-of-the-art RL fine-tuned diffusion model.

\end{abstract}

\section{Introduction}
Inverse protein folding is the task of designing amino acid sequences that fold into a target protein backbone. Discrete diffusion models have emerged as a leading architecture for this task \citep{inverse_folding_old, gruver2023proteindesignguideddiscrete, cemri2024discrete, wang2025finetuningdiscretediffusionmodels}. While these models can successfully generate sequences that fold into target backbones, scientists may also target additional attributes such as stability \citep{wang2025finetuningdiscretediffusionmodels} or $\beta$-sheets \citep{gruver2023proteindesignguideddiscrete}. Many sequences can fold into similar structures, yet only a few may satisfy target criteria \citep{proteinguide}. Given an arbitrary protein model, our goal is to generate proteins with desirable attributes by efficiently using a reward model that quantifies those attributes. Test-time alignment is a category of methods that use additional compute during inference to generate desirable samples. Existing test-time alignment of discrete diffusion models can be organized into three groups: token logit alignment, tree-search, and re-noising feedback. Token logit alignment skews probabilities of tokens during the reverse process such as in \citep{proteinguide} and \citep{gruver2023proteindesignguideddiscrete}. Tree-search techniques, like those discussed in \citep{huang2025bestofnbestthemcoverage, Darmawan2025.09.14.676087, uehara2025inferencetimealignmentdiffusionmodels}, explore different denoising trajectories in the reverse process and select the highest-reward sample. We focus on the third and less studied group of methods: re-noising feedback. This method, discussed in \citet{remasking_diffusion}, naturally exploits the masking structure of discrete diffusion models in the following sequence: generate a candidate sequence, re-mask a subset $S \subseteq [L]$ of positions to edit, and re-sample those positions. This type of method uniquely treats the model as a black box instead of modifying its denoising trajectory. Previous sequence editing methods have used independent random re-masking; however, the actual alignment comes from other techniques such as importance sampling or classifier-free guidance rather than solely re-masking \citep{lee2025effectivetesttimescalingdiscrete, reid2022diffuserdiscretediffusioneditbased, remasking_diffusion, gruver2023proteindesignguideddiscrete}. The addition and removal of noise also have parallels with continuous diffusion model image editing methods, which transform images into partially noisy latents and then re-sample a segment of the diffusion reverse process \citep{p_to_p_img_edit, text_inversion_img_edit, sde_img_edit}. Similar to the discrete case, alignment in these methods comes from techniques other than noise generation.

We claim that re-masking without independence assumptions can be an effective alignment method on its own. This approach is useful because its modularity allows any other alignment method to be improved within a re-masking feedback loop. Formally, we focus on selecting an edit-set $S$ that maximizes a value function $f$ which relates edit-sets to alignment rewards. For an arbitrary value function $f: 2^{[L]} \to \mathbb{R}$, maximization over $2^L$ candidates is intractable, which supports the use of independence assumptions in past works. Our work identifies structure in edit-set value functions which enables approximations that avoid combinatorial challenges during learning and optimization.

\begin{figure}[t]
        \centering
        \includegraphics[width=0.9\linewidth]{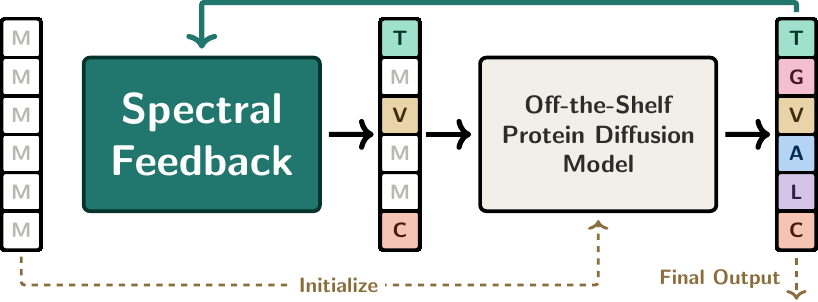}
        \vspace{-3pt}
    \caption{Spectral Feedback loop using targeted edits to iteratively improve attributes of a protein sequence with discrete diffusion.}
    \label{fig:fig1}
    \vspace{-20pt}
\end{figure}
 
\begin{wrapfigure}{r}{0.5\textwidth}
  \centering
  \vspace{-5pt}
  \includegraphics[width=\linewidth]{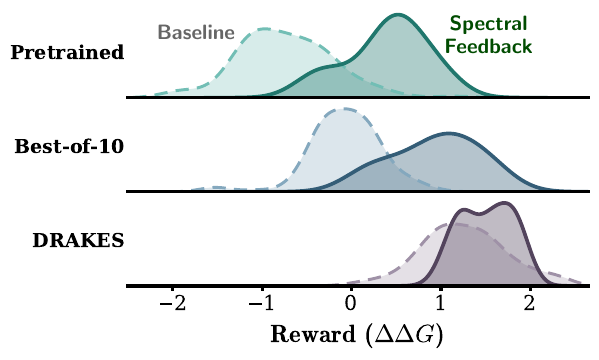} 
  \caption{Spectral Feedback in different protein model inference settings, aligning for stability ($\Delta \Delta G$ - kcal/mol). These are reward evaluation distributions from experiments in Section \ref{sec:experiments}.}
  \label{fig:halfpage-right}
  \vspace{-5pt}
\end{wrapfigure}
 
The expansion of $f$ in the Fourier basis assigns every subset $T \subseteq [L]$ a coefficient $F(T)$ that measures the strength of the interaction among positions in $T$ \citep{odonnell2014analysis, spright, butler2025proxyspexinferenceefficientinterpretabilitysparse}. A sparse Fourier expansion contains a small number of coefficients that capture most of the variation in $f$. Sparsity of this form is consistent with the epistatic structure of biological sequences \citep{epistasis_pattern, epistatic_net, crispr}, in which the joint effect of mutating multiple amino acids is dominated by a small number of interacting groups. Because the interactions that determine the physical properties of a protein are sparse, we expect that the interactions between edits of these very same amino acids will also reflect a sparse structure. We verify this empirically for protein sequence edit-set value functions.

Importantly, Fourier sparsity makes it tractable to approximate $f$ from a limited number of queries. Instead of requiring $2^L$ reward evaluations, the number of samples needed for sparse Fourier recovery grows only with the sparsity level \citep{7042831, spright}, which we find to be small in practice for edit-set value functions \citep{kang2025spexscalingfeatureinteraction, butler2025proxyspexinferenceefficientinterpretabilitysparse}. We also find that most spectral energy is concentrated in lower-order terms, with some value function instances being nearly linear.

Motivated by our observations of sparsity, we introduce \textbf{Spectral Feedback}, a feedback-based test-time alignment method for discrete diffusion. Our method leverages sparse Fourier recovery to approximate the combinatorial edit-selection problem in a tractable form. Spectral Feedback treats the underlying diffusion model as a black box: it requires only the ability to sample from the model conditioned on a partially masked input, and never modifies its weights or its denoising trajectory. This makes it compatible with any pre-trained discrete diffusion model and composable with other alignment techniques. We apply Spectral Feedback on top of a pre-trained model, Best-of-N sampling, and an RL fine-tuned model (DRAKES \citep{wang2025finetuningdiscretediffusionmodels}). 

We evaluate Spectral Feedback on protein backbones from the Megascale dataset \citep{tsuboyama2023}, aligning sequences to a stability oracle that predicts $\Delta\Delta G$, the difference in Gibbs free energy between a generated sequence and its wild-type variant. To detect reward over-optimization, we separately track self-consistency RMSD ($scRMSD$) between the target backbone and the ESMFold-predicted structure of each sample \citep{esm_fold} as well as naturalness with ProtGPT2 log-likelihoods \citep{ferruz_protgpt2_2022}. 

\paragraph{Contributions.}
\begin{enumerate}
    \item \textbf{Formulation.} 
        We formalize a feedback loop that relates the selection of edit-positions at each iteration to an alignment reward oracle by defining an edit-set value function. The test-time alignment problem is then to maximize this intermediate value function rather than the reward oracle directly. This creates a combinatorial challenge that prior refinement methods largely avoid, since the value of editing one position can depend on which other positions are edited simultaneously.
    \item \textbf{Sparsity analysis.} We characterize the Fourier spectra and the sparsity structure of edit-set value functions across protein targets. Our analysis suggests that some proteins are dominated by first-order effects, while others involve high-order interactions, giving an empirical explanation for performance differences between simpler position-wise edit methods and high-order sparse recovery methods.

    \item \textbf{Method.} We introduce Spectral Feedback, which learns a sparse Fourier approximation of an edit-set value function, and then uses the approximation to efficiently identify promising edits at each iteration of a diffusion feedback loop. Spectral Feedback is model-agnostic, which allows it to improve the performance of any pre-trained, test-time aligned, or fine-tuned model.
    \item \textbf{Empirical results.} On inverse folding of protein backbones in the Megascale dataset, Spectral Feedback improves alignment rewards for several different underlying protein models. Compared with several greedy or gradient-based baseline edit-set selection methods, Spectral Feedback offers a favorable reward–latency trade-off.
\end{enumerate}

\section{Preliminaries}
\textbf{Discrete Diffusion Models.} We establish the notation here, while a complete treatment appears in Appendix~\ref{app:appendixa}. Consider a vocabulary $V$, sequence length $L$, and sample space $\mathcal{X}:=V^L$. We consider a masked discrete diffusion model whose forward process progressively masks tokens via a Continuous Time Markov Chain (CTMC) with rate matrices $Q_t$. Its probability mass vector evolves according to $dp_t/dt = Q_t p_t$, with $p_0 = p_{\text{data}}$, where $Q_t(y,x)$ denotes the transition rate from $x$ to $y$  \citep{liang2025discretediffusionmodelsnovel}. The reverse process, initialized from a fully masked sequence $x_T$, progressively unmasks tokens by learning a score function $p_t(y)/p_t(x)$ from training data. A Hamming distance constraint ($Q_t(x,y)=0$ for $d(x,y)>1$) restricts each transition to a single-token flip. Practical samplers can update multiple tokens per time step by approximating their transitions as conditionally independent given the current sequence.

\textbf{Feedback with Edit-Positions.} Given a protein sequence $x$ of length $L$, feedback selects a subset of positions to re-mask and then re-samples them from the diffusion model. Let $S\subseteq [L]$ denote a set of edit-positions. Applying $S$ constructs a partially masked sequence $\tilde{x}$:
\begin{align}
\label{eq:remask}
    \tilde{x}_i=\left\{\begin{aligned}
        &MASK \; &\text{if }i\in S\\
        &x_i \; &\text{if }i\notin S
    \end{aligned}\right.
\end{align}
A new sequence $z\sim p_{\text{pre}}(\cdot | \tilde{x})$ is generated by running the reverse process initialized with $\tilde{x}$. The refinement scheme alternates between selecting $S$ and re-sampling, seeking edit-positions that maximize a value function: $S^*=\underset{S\subseteq [L], |S|\leq k}{\arg \max} \;f(S)$. The cardinality bound $k$ balances alignment performance with the computational cost of solving the maximization problem and re-sampling at the edit positions. Prior methods select edit-positions randomly or independently \citep{lee2025effectivetesttimescalingdiscrete, reid2022diffuserdiscretediffusioneditbased, gruver2023proteindesignguideddiscrete}. Our method instead searches for optimal \emph{sets} of positions by exploiting their joint reward structure, as described in Section~\ref{sec:spectral_feedback}. 

We define the total number of steps $N$ of a feedback alignment process as the total number of proteins generated during the process. Our algorithm generates an initial protein sequence before executing the feedback loop, so it has $N - 1$ feedback iterations. At each step, we initialize the reverse process with $\tilde{x}$. We then sample $z$ by running the reverse process from the first time-step to completion. In future work, it could be natural and more efficient to use a later time-step that relates to the masking pattern of $\tilde{x}$ and the noise schedule of the diffusion model. 

\section{Related Work}

\textbf{Protein Sequence Design and Inverse Folding.}
Recent work has produced strong generative models for inverse folding. ProteinMPNN \citep{proteinmpnn} introduced a graph neural network conditioned on backbone geometry that autoregressively generates sequences. Discrete diffusion variants extend the structured methods to non-autoregressive masked generation \citep{inverse_folding_old, gruver2023proteindesignguideddiscrete, cemri2024discrete, wang2025finetuningdiscretediffusionmodels}. DRAKES \citep{wang2025finetuningdiscretediffusionmodels} fine-tunes a discrete diffusion inverse folding model with a differentiable reward signal, improving alignment to a stability oracle at the cost of model-specific training.

\textbf{Inference-Time Alignment for Discrete Generative Models.}
Inference-time alignment methods spend additional compute at sampling time to improve reward without retraining. Best-of-N draws $N$ independent samples and returns the highest-reward one \citep{huang2025bestofnbestthemcoverage}; it is the dominant approach in protein design due to its simplicity \citep{enzyme_design, antibody_design, Darmawan2025.09.14.676087}, but its reward improvement grows slowly with $N$. Beam Search retains only the highest-reward partial trajectories at each reverse-process step \citep{uehara2025inferencetimealignmentdiffusionmodels}. Token-level guidance methods modify the score function during the reverse process: ProteinGuide uses Bayesian conditioning with classifiers trained on partially masked sequences \citep{proteinguide}, and NOS performs Langevin updates in the protein embedding space \citep{gruver2023proteindesignguideddiscrete}.

\textbf{Sequence and Image Editing Methods.}
Sequence editing techniques often progressively transform an existing sample toward a target objective. \citet{lee2025effectivetesttimescalingdiscrete} use Multiple-Try Metropolis with uniform random noise to refine intermediate reverse-process time-steps. \citet{remasking_diffusion} design re-masking schedules that facilitate better sample quality and can benefit existing reward guidance methods, but do not themselves provide reward guidance. \citet{gruver2023proteindesignguideddiscrete} apply embedding-space Langevin updates at positions sampled with probabilities proportional to reward-gradient magnitudes, and DiffusER uses random Levenshtein edits as the forward process \citep{reid2022diffuserdiscretediffusioneditbased}. In the language model setting, Self-Refine \citep{madaan2023selfrefineiterativerefinementselffeedback} and multi-agent pipeline optimization \citep{xue2025improveiterativemodelpipeline} use LLM-generated feedback rather than reward oracles. 

In parallel to sequence editing, several image editing methods re-noise and re-sample image latents using the continuous diffusion forward and reverse processes which involve independent Gaussian noise \citep{p_to_p_img_edit, text_inversion_img_edit, sde_img_edit}. A common pattern across both discrete and continuous diffusion editing methods is that noise is generated independently. The discrete case offers a unique opportunity for targeted edit selection since the noise is made up of masks that directly correspond to individual tokens. Spectral Feedback uses this structure during test-time alignment by selecting tokens to re-mask and re-sample.

\textbf{Sparse Boolean Function Recovery and Epistasis.}
The combinatorial structure of edit-set selection connects to a long line of work on sparse Boolean function recovery. Epistasis, in which genes or amino acids interact to determine biological traits \citep{epistasis_def}, is documented to exhibit sparse high-order structure \citep{epistasis_pattern, epistatic_net, crispr}, and similar sparsity patterns have been observed in language and vision data \citep{butler2025proxyspexinferenceefficientinterpretabilitysparse}. LASSO recovers sparse linear models efficiently \citep{lasso} but requires explicit enumeration of interaction features, which scales exponentially with order. Recent algorithms \citep{7042831, spright,amrollahi2019efficiently,kang2025spexscalingfeatureinteraction,butler2025proxyspexinferenceefficientinterpretabilitysparse} address this by recovering high-order Fourier coefficients with sub-exponential sample complexity. We use this line of work to approximate edit-set value functions and to study how the sparsity structure of these functions varies across proteins.

\section{Spectral Feedback}\label{sec:spectral_feedback}

We introduce Spectral Feedback, a feedback loop that iteratively selects and applies edits to a sequence with a discrete diffusion model. Our algorithm leverages sparse interactions in edit-set value functions to efficiently select promising edits. In this section, we study these sparse interactions and then describe the details of our algorithm.
\subsection{Spectral Function Approximation}
\label{sec:spectral_approximation}
 \begin{center}
\begin{figure}[h]
\vspace{-10pt}
    \centering
\begin{minipage}{\textwidth}
    \centering
    \includegraphics[width=0.45\textwidth]{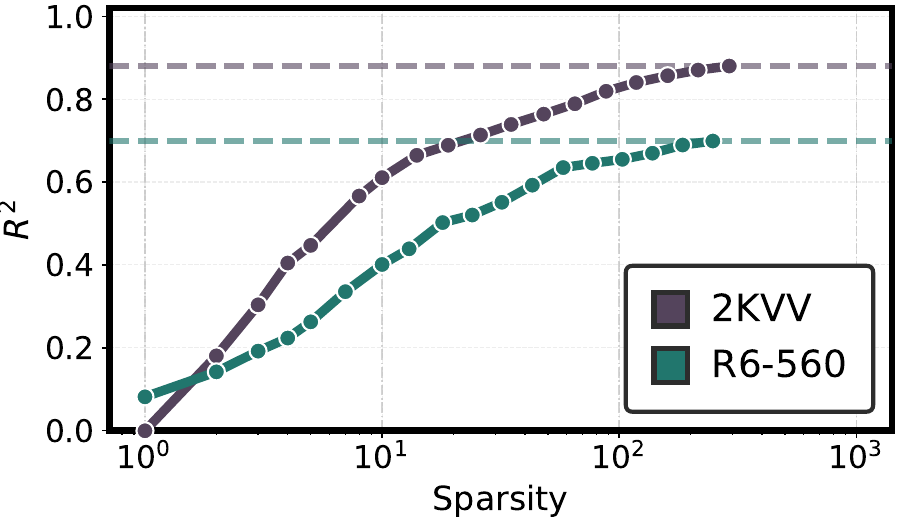}
    \hspace{0.5cm}
\includegraphics[width=0.45\linewidth]{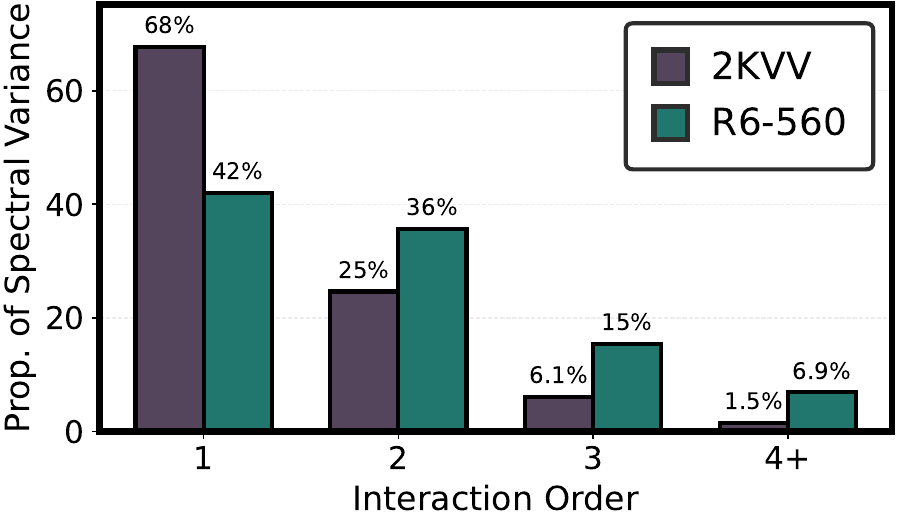}
\end{minipage}
\vspace{-5pt}
\caption{Spectral analysis of the value function $f_{avg}$ with the pre-trained diffusion model. The plots show results for two generated protein sequences conditioned on the 2KVV and R6-560 (r6\_560\_TrROS\_Hall) backbones from the Megascale dataset. Experimental details and additional plots are provided in Appendix~\ref{sec:r2_exps}.}
\label{fig:sparsity}
\vspace{-20pt}
\end{figure}
\end{center}

The goal of edit-set selection is, given a protein sequence $x$, target backbone $y$, and protein model $p(\cdot |S,x,y)$, to find a set $S^*$ that maximizes a value function $f:2^{[L]}\rightarrow \mathbb{R}$. In terms of edits, this value function should reward more promising edit selections (i.e., those with higher expected rewards after re-sampling). Given a reward oracle $r(\cdot)$, we use the following value functions depending on the sampling techniques used for the underlying protein model:

\vspace{-15pt}
\begin{align}
\label{eq:f_tilde}
f_{avg}(S, x_t, y)=\frac{1}{n}\sum_{i=1}^nr(z_i), \quad f_{max}(S,x_t,y)=\max \{r(z_1),\ldots,r(z_n)\}
\end{align}
where $z_i\sim p(x_{t-1}|S,x_t,y \text{ and $x_{t-1}$ is unmasked})$ for a discrete diffusion model $p$.
Other value functions exist---such as trained predictors for partially masked sequences \citep{gruver2023proteindesignguideddiscrete, proteinguide} or posterior mean approximations \citep{uehara2025inferencetimealignmentdiffusionmodels}---though we selected this method for computational feasibility. Still, we find that this single-step approximation of the reverse process leads to strong alignment performance.

A value function $f$ admits a Fourier transform $F : 2^{[L]} \rightarrow \mathbb{R}$ as follows \citep{odonnell2014analysis}:

\vspace{-13pt}
 \begin{align}
     \text{Transform: }F(T)=\frac{1}{2^L}\sum_{S\subseteq [L]}(-1)^{|S\cap T|}f(S), \quad \text{Inverse: }f(S)=\sum_{T\subseteq [L]}(-1)^{|T\cap S|}F(T).
 \end{align}
 \vspace{-7pt}

While there are $2^L$ coefficients that fully describe $f$, if we can use an approximation $\hat{f}\approx f$ with significantly fewer coefficients, then learning $\hat{f}$ can become tractable. The sparse support also allows for optimizing via a small integer program. Parseval's theorem gives us
\begin{align}
    \frac{1}{2^L}\sum_{S\subseteq [L]} \left(f(S) - \hat{f}(S)\right)^2 = \sum_{T\subseteq [L]} \left(F(T) - \hat{F}(T)\right)^2. 
\end{align}
Consequently, if $f$ admits a \emph{sparse} Fourier representation---i.e., most of its energy is concentrated in a small number of coefficients---then accurately recovering only those dominant coefficients suffices to approximate $f$ well. This motivates our use of sparse recovery methods to estimate $F$ from a limited number of evaluations of $f$.

\vspace{-15pt}
\begin{align}
    R^2=1-\frac{||\hat{f}-f||^2}{||f-\bar{f}||^2}, \quad \text{where }||f||^2=\sum_{S\subseteq [L]}f(S)^2, \bar{f}=\frac{1}{2^L}\sum_{S\subseteq [L]}f(S).
\end{align}
\vspace{-15pt}

We use the sparse recovery algorithm SPEX \citep{kang2025spexscalingfeatureinteraction} to accurately compute Fourier coefficients that describe a chosen edit-set value function during a single feedback iteration. Figure~\ref{fig:sparsity} shows results from this spectral analysis of $f_{avg}$ for two distinct protein backbones. Most spectral variance is captured by coefficients with order at most three and value function approximations for both proteins reach an $R^2$ of at least 0.7. These results indicate that the Fourier coefficients representing $f_{avg}$ are sparse and concentrated at low orders. Similar results are shown in Appendix~\ref{sec:r2_exps} for the whole test set with $f_{avg}$ and $f_{max}$. 

An interesting difference between the two held-out protein examples is that first-order coefficients dominate for \textit{2KVV} while high-order interactions are more significant for \textit{R6-560}. These differences suggest that interactions between edit-positions depend on the properties of the underlying proteins, which are strongly influenced by interactions among their respective amino acids. When mainly first-order coefficients describe the value function, like for \textit{2KVV}, a first-order sparse recovery method will likely be sufficient to learn a useful $\hat{f}$. We further investigate this in Appendix~\ref{app:spec_study}.
 
\subsection{Spectral Feedback Algorithm}
\label{sec:spec_feedback_alg}
\begin{figure}[H]
    \vspace{-15pt}    
    \centering
    \includegraphics[width=1\linewidth]{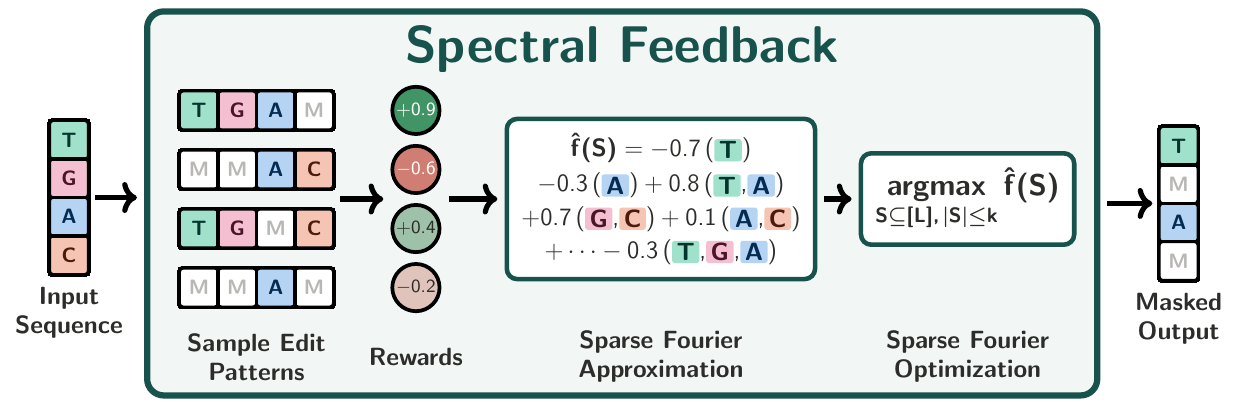}
    \caption{Spectral Feedback learns a sparse Fourier representation of the edit-set value function, then solves for the max-reward set via integer optimization over the sparse coefficients.}
    \label{fig:spec_feedback_vis}
    \vspace{-10pt}
\end{figure}

The Spectral Feedback algorithm alternates edit-position selection and model sampling, using any off-the-shelf masked diffusion model as shown in Figure~\ref{fig:fig1}. The algorithm is described below.

\textbf{Step 1 - Select edit-positions (Algorithm \ref{alg:spectral}).} Given an amino acid sequence $x$ and an edit budget $k$, the first step selects a set of edit-positions $S\subseteq [L]$ such that $|S|\leq k$. This step, depicted in Figure~\ref{fig:spec_feedback_vis}, is done in the following sub-steps: sample sets of edit-positions, evaluate corresponding rewards, fit a sparse Fourier approximation, and solve for the optimal positions. The sample sets are assigned rewards using Equation \ref{eq:f_tilde}. A sparse Fourier approximation is then fit with the sampled set-reward pairs. This can be done with any sparse recovery method. Finally, edit-positions are selected by solving an optimization problem over the learned sparse Fourier approximation. The setup of this optimization problem is described in Appendix~\ref{supp:optimization}.

\textbf{Step 2 - Sample the Protein Model (Algorithm \ref{alg:spectral_feedback}).} After finding a set $S$ of edit-positions, the corresponding amino acids at each position are re-masked as described in Equation \ref{eq:remask}. The partially masked sequence is then passed back through the discrete diffusion reverse process.

\textbf{Sparse Recovery Method}.
While any sparse recovery method can be used in Spectral Feedback (e.g., \citet{kang2025spexscalingfeatureinteraction, spright,amrollahi2019efficiently}), we use ProxySPEX \citep{butler2025proxyspexinferenceefficientinterpretabilitysparse}, a method that uses Gradient Boosted Trees to learn a sparse Fourier representation due to its sample efficiency. To sample edit-sets for ProxySPEX, we randomly generate binary masks in which each element is an independent Bernoulli random variable that indicates inclusion in the set.

\begin{figure*}[!t]
\centering

\textbf{Require:} $L,D,k,N\in\mathbb{N}$, $0<\gamma<1$. Vocabulary $V$ with mask token $M$. Protein backbone $y\in Y$.

\vspace{-1.5em}

\begin{minipage}[t]{0.48\textwidth}
\begin{algorithm}[H]
\caption{Spectral Edit-Set Selection}
\label{alg:spectral}
\begin{algorithmic}
\State \textbf{Input:} $x \in V^L$
\State{$S_j \subseteq [L]\; \forall j \in [D]$}
\State{$i\in S_j \;\text{with prob.} \;\gamma$} \Comment{Sample edit sets}
\State{$\tilde{r}_j \gets f(S_j,x,y)$} \Comment{Equation \ref{eq:f_tilde}}
\State{$\hat{f} \gets \text{ProxySPEX}(\tilde{r},S)$} \Comment{Sparse Recovery}
\State{$S^* \gets \displaystyle \underset{S \subseteq [L], \;|S|\leq k}{\arg \max}\; \hat{f}(S)$} \Comment{Appendix~\ref{supp:optimization}}
\State{$\textbf{return} \; S^*$}
\end{algorithmic}
\end{algorithm}
\end{minipage}
\hfill
\begin{minipage}[t]{0.48\textwidth}
\begin{algorithm}[H]
\caption{Protein Model Feedback}
\label{alg:spectral_feedback}
\begin{algorithmic}
\State \textbf{Input:} $x \in V^L$
\State{$x_i \gets M \; \forall i\in[L]$} \Comment{Full mask initialization}
\State{$S\gets [L]$} \Comment{Edit all tokens initially}
\For{$i$ in $1,\ldots,N-1$}
    \State{$x\sim p(\cdot| S,x,y)$} \Comment{Protein model}
    \State{$S\gets \text{EditSelection}(x)$} \Comment{Algorithm \ref{alg:spectral}}
\EndFor
\State{$x\sim p(\cdot| S,x,y)$} \Comment{Last feedback iteration}
\State{$\textbf{return} \; x$}
\end{algorithmic}
\end{algorithm}
\end{minipage}
\vspace{-10pt}
\end{figure*}

\FloatBarrier

\textbf{Value Function}.
The choice of the value function from Equation \ref{eq:f_tilde} depends on how the underlying diffusion model is sampled. When Spectral Feedback is applied directly to a diffusion model such as a pre-trained or RL fine-tuned model like DRAKES, we use $f_{avg}$ to approximate the expected reward of the resulting proteins. However, additional test-time alignment methods can be applied on top of a diffusion model within the feedback loop. We demonstrate this by applying Spectral Feedback to Best-of-N with a pre-trained diffusion model. To more naturally follow the dynamics of Best-of-N, we use $f_{max}$ for the edit-set value function. The value function $f_{max}$ should be used instead of $f_{avg}$ whenever there is an underlying reward-maximizing test-time alignment method, such as Best-of-N. In these methods, only the highest-reward sample matters, so we should not penalize an edit-set for low-reward samples by using an average.

\section{Experiments}
\label{sec:experiments}

\begin{figure}[t]
    \centering
    \includegraphics[width=\linewidth]{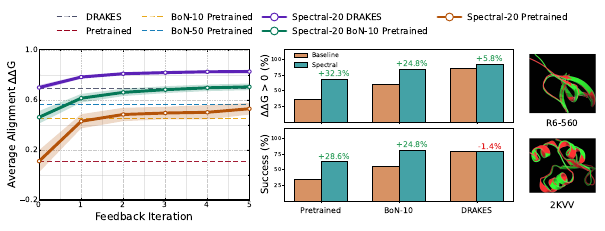}
    \vspace{-20pt}
    \caption{\textbf{(left)} Spectral Feedback with $k=20$ is applied to several underlying protein models: pre-trained, Best-of-10 with pre-trained, and DRAKES. Spectral Feedback improves the alignment reward of each algorithm in only a few iterations. \textbf{(middle)} Percentage of stable proteins (evaluation $\Delta \Delta G >0$) and the resulting success rate when $scRMSD<2$ are shown. The $scRMSD$ constraint and metric averages are in Table~\ref{fig:spec_feedback_modular_table}. (\textbf{right}) ESMFold protein structures comparing the wild-type (\textit{red}) and Spectral Feedback (\textit{green}) sequences demonstrate that Spectral Feedback preserves the underlying inverse protein folding model's capabilities.}
    \vspace{-5pt}
    \label{fig:spec_feedback_modular_plots}
    \vspace{-10pt}
\end{figure}

\subsection{Experimental Setup}
We use protein backbones from the Megascale dataset to evaluate our algorithm. The evaluation metrics are \textit{stability}, $scRMSD$, and \textit{naturalness}.
\begin{itemize}
    \item The \textit{stability} oracle represents the quantity $\Delta \Delta G=\Delta G_{wild}-\Delta G_{align}$ where $\Delta G$ is the Gibbs free energy of the corresponding protein. A positive $\Delta\Delta G$ means a lower $\Delta G$ than the corresponding wild-type sequence, and thus a more stable protein.

    \item The self-consistency RMSD ($scRMSD$) oracle represents the RMSD between the target protein backbone of the inverse folding process and the predicted folded structure of a generated amino acid sequence. We use ESMFold \citep{esm_fold} to predict the structure of a sequence and then calculate the RMSD with the given backbone.

    \item The \textit{naturalness} oracle represents the log-likelihood of an amino acid sequence. We use ProtGPT2 \citep{ferruz_protgpt2_2022}, an autoregressive language model that captures the distribution of protein sequences. This metric complements $scRMSD$ in identifying overfitting to the stability oracle.
\end{itemize}
We use the $\Delta \Delta G$ metric as the primary target reward for aligning the underlying protein model, though we also include results for alignment with the ProtGPT2 oracle and a multi-objective oracle in Appendix~\ref{app:additional_alignment_oracles}. DRAKES trained two $\Delta \Delta G$ oracles with the Megascale dataset: one for alignment and one for evaluation \citep{wang2025finetuningdiscretediffusionmodels}. To fairly compare with their work, we also align and evaluate $\Delta\Delta G$ with these oracles. We also use the same validation/test dataset split. Following previous works \citep{campbell2024generativeflowsdiscretestatespaces,nisonoff2025unlockingguidancediscretestatespace,wang2025finetuningdiscretediffusionmodels}, we classify a \textit{successful} protein as one that satisfies $\Delta\Delta G>0$ and $scRMSD < 2$. These constraints ensure that proteins are at least as stable as their wild-type variants and that their folded structure resembles the target structure. 

We use both the pre-trained and RL models from DRAKES to test our method. These models are built with the ProteinMPNN architecture which uses a graph neural network to capture local interactions between amino acids and incorporates extracted features from the backbone structure \citep{proteinmpnn, wang2025finetuningdiscretediffusionmodels}. We also test Spectral Feedback alignment of Best-of-N with the pre-trained model. Following our discussion in Section \ref{sec:spec_feedback_alg}, we use $f_{avg}$ for the pre-trained and DRAKES models, and $f_{max}$ for the pre-trained model with Best-of-N.

\subsection{Baseline Edit Selection Methods}
Spectral Feedback is unique in that it chooses edit-positions directly based on interactions between tokens. To establish that these interactions are important, we compare the algorithm's performance with methods that do not model interactions. These are described below.

\textbf{Gradient-Weighting}. This method selects edit-positions based on the reward function's gradient at those positions. In particular, for a sequence $x$, the $i^{th}$ token position is assigned a weight $w(i)= h_i(x)^T\nabla_i r(h_i(x))$ where $h_i(x)$ is the embedding for the $i^{th}$ token. This method is meant to mirror the use of saliency maps in \citet{gruver2023proteindesignguideddiscrete} for edit-position selection in Langevin dynamics updates, but with signed importance scores. For edit-positions, we select the tokens with the most negative scores since these correspond to tokens that lead to lower reward.

\textbf{Token Exclusion}. Whereas Spectral Feedback directly selects an optimal set of edit-positions, Token Exclusion looks at individual tokens to determine edit-positions. It calculates the value function when masking each token in isolation and then selects the highest-value positions.

\textbf{Argmax Selection}. To show that the sparse recovery methods successfully use observed edit sets to identify a better edit set, we compare against the argmax method which selects the best observed sample edit set. While this method will suffer if sample edit-sets are suboptimal, a sparse recovery method still has hope of finding a rare but optimal edit-set.

\textbf{Hill Climbing}. We implement a randomized 1-flip local search, as described in \citet{hillclimb}, for an alternative greedy baseline. This method iteratively improves its edit selection by uniformly randomly adding or removing an edit-position from the current selection and accepting the change if it improves the reward. A drawback is that the algorithm is inherently serial so parallelization is limited to single-set reward calculations.

\vspace{-5pt}
\subsection{Results on Protein Inverse Folding}

\begin{table}[h]
    \centering
    \resizebox{\textwidth}{!}{
    \begin{tabular}{l*{5}{c}}
    \toprule
    Method 
        & Align $\Delta\Delta G$ $\uparrow$ 
        & Eval $\Delta\Delta G$ (avg) $\uparrow$
        & scRMSD (avg) $\downarrow$ 
        & scRMSD $< 2$ (\%) $\uparrow$ 
        & Log-Likelihood $\uparrow$ \\
    \midrule

    pre-trained
        & 0.10
        & -0.49
        & 1.10
        & 91.8
        & -147.8 \\
        
    \rowcolor{gray!15}
    + \textbf{Spectral}
        & \textbf{0.53}\gd{+0.43}
        & \textbf{0.33}\gd{+0.82}
        & \textbf{1.06}\gd{-0.04}
        & \textbf{94.2}\gd{+2.4}
        & \textbf{-146.5}\gd{+1.3} \\
        
    \midrule

    BoN-10
        & 0.45
        & 0.09
        & 1.09
        & 92.4
        & -148.1 \\
        
    \rowcolor{gray!15}
    + \textbf{Spectral}
        & \textbf{0.70}\gd{+0.25}
        & \textbf{0.86}\gd{+0.77}
        & \textbf{1.07}\gd{-0.02}
        & \textbf{96.7}\gd{+4.3}
        & \textbf{-147.5}\gd{+0.6} \\
        
    \midrule

    DRAKES
        & 0.69
        & 0.99
        & \textbf{1.17}
        & \textbf{93.4}
        & -157.5 \\
        
    \rowcolor{gray!15}
    + \textbf{Spectral}
        & \textbf{0.83}\gd{+0.14}
        & \textbf{1.38}\gd{+0.39}
        & 1.53\bd{+0.36}
        & 85.8\bd{-7.6}
        & \textbf{-157.7}\bd{-0.2} \\
        
    \bottomrule
\end{tabular}
    }
    \vspace{5pt}
    \caption{Spectral Feedback improves alignment and evaluation oracle rewards across all three base methods, with the largest gains on the pre-trained and Best-of-10 baselines. DRAKES + Spectral achieves the highest evaluation $\Delta\Delta G$ overall (1.38) but trades off structural quality ($scRMSD$). Each pair shows baseline followed by its Spectral-enhanced counterpart, using $k=20$.} 
    \label{fig:spec_feedback_modular_table}
    \vspace{-20pt}
\end{table}

\textbf{Modular Integrations.}
We first evaluate Spectral Feedback's performance as a modular alignment technique. The reward trajectories in Figure~\ref{fig:spec_feedback_modular_plots} are of the alignment oracle whereas the bar plot shows the evaluation oracle results. The evaluation oracle distributions are also visualized in Figure~\ref{fig:halfpage-right}. These results show that Spectral Feedback improves the performance of whichever underlying protein model it samples from. In particular, in Table~\ref{fig:spec_feedback_modular_table}, the reward increases for both the alignment and evaluation oracles. The relatively stable log-likelihoods and the high success rates in Figure~\ref{fig:spec_feedback_modular_plots} further support Spectral Feedback's robustness. However, the average $scRMSD$ increases when Spectral Feedback is applied to DRAKES, thus limiting the gains that can be made in protein success rate. This trade-off is consistent with the reward–structure trade-off already present in high-stability regions within the DRAKES protein distribution. This trade-off is discussed in detail in Appendix~\ref{sec:scrmsd_drakes_disc}.

Figure~\ref{fig:spec_feedback_modular_plots} shows that Spectral Feedback surpasses Best-of-10’s alignment reward after two feedback iterations,  though not at a lower computational cost: edit-position selection requires many diffusion model calls to evaluate Equation~\ref{eq:f_tilde}. A lightweight learned value model, as used in \citet{gruver2023proteindesignguideddiscrete} and \citet{proteinguide}, could reduce this overhead. Despite this cost, Spectral Feedback improves on Best-of-N beyond its diminishing returns. A detailed discussion of computational costs and scaling beyond diminishing returns in Best-of-N is provided in Appendix~\ref{app:compute_complexity}.

Many existing alignment methods target either an argmax or a weighted Gibbs distribution, $p_a(x) \propto p_{pre}(x)\exp(\alpha r(x))$. Techniques such as Best-of-N, Beam Search, and MCTS all directly approximate an argmax by searching the sequence space for the highest-reward sample. Common RL objectives with KL regularization, like the objective used in DRAKES \citep{wang2025finetuningdiscretediffusionmodels}, are maximized by weighted Gibbs distributions. Our experiments suggest that Spectral Feedback could improve additional alignment methods because of the similarity of the target distributions across alignment techniques and the modularity of the feedback mechanism.

\textbf{Baseline Edit-Position Selection}. We compare Spectral Feedback with baseline edit-set selection methods, evaluating how alignment reward scales with the number of edit-set samples and total feedback iteration time. All results are from a single feedback iteration. Exclusion and Gradient are faster but achieve lower rewards than Spectral Feedback, Argmax Selection, and Hill Climbing. Spectral Feedback and Hill Climbing exhibit similar sample-scaling, but Spectral Feedback achieves a better latency–reward trade-off because Hill Climbing’s sequential search limits batching across edit-set samples. The batched sampling and reward oracle evaluation required by Equation~\ref{eq:f_tilde} limit the number of edit-set values that Spectral Feedback and Argmax Selection can compute in parallel. A lightweight value function that avoids these operations could support larger batches which would increase the computational advantage over the sequential search in Hill Climbing. Argmax Selection and Spectral Feedback use identical samples in these experiments, but Spectral Feedback incurs additional overhead from ProxySPEX fitting and optimization. Still, the latency trade-off plot shows that this overhead is insignificant as the number of samples increases.

\begin{figure}[ht]
    \centering
    \includegraphics[width=\linewidth]{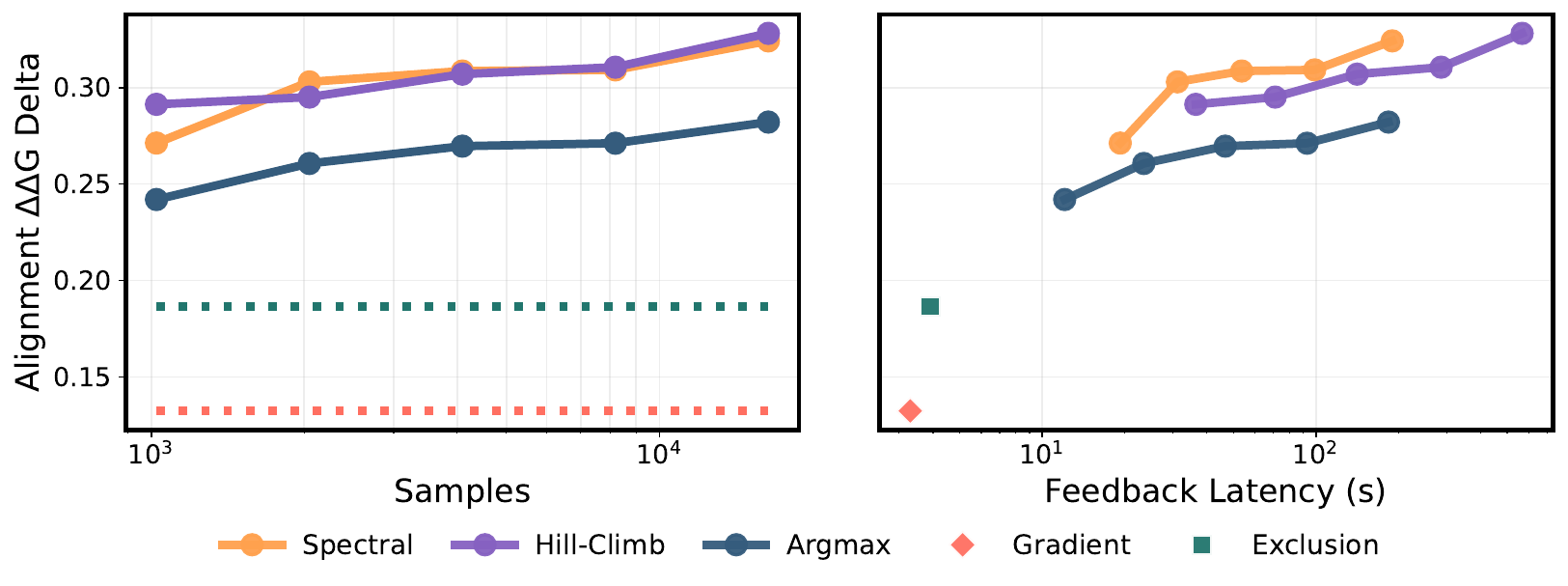}
    \vspace{-20pt}
    \caption{Spectral Feedback shows strong performance in scaling of edit-set samples and latency.}
    \label{fig:baseline_edit_position_selection_curve}
    \vspace{-5pt}
\end{figure}
\newpage
\section{Discussion}
\textbf{Conclusion}.
In this work, we study test-time alignment of discrete diffusion models for inverse folding with access to protein sequence reward oracles. We introduce Spectral Feedback, an algorithm that iteratively identifies a promising subset of edit-positions for a generated sequence and re-samples the corresponding amino acids through the diffusion reverse process. The value of editing an amino acid depends on which other amino acids are edited alongside it, making edit-set selection a challenging combinatorial problem. We address this challenge by empirically showing that edit-set value functions can have sparse Fourier representations. We use sparse recovery methods to approximate these functions and then optimize the approximations to select promising edit-sets. On the Megascale dataset, Spectral Feedback improves stability alignment rewards across three settings: the pre-trained model, Best-of-N, and DRAKES. It also preserves structural self-consistency and increases protein success rates for the pre-trained and Best-of-N models, while applying Spectral Feedback to DRAKES introduces a stability–structure trade-off.

\textbf{Limitations \& Future Work}. A limitation of our sparse recovery method is that high global $R^2$ does not necessarily correspond to accurate argmax recovery. We explore using an exponentially tilted objective to address this in Appendix~\ref{app:exp_tilt_val_func}. Another sparse recovery limitation is that a large number of edit-set samples is required to successfully learn a sparse Fourier representation. However, once in this high-sample regime, our algorithm scales better than competing methods. In the future, the scaling of alternative sparse recovery methods such as SPEX \citep{kang2025spexscalingfeatureinteraction} could be explored.

\begin{ack}
 We thank Jennifer Listgarten and her lab group for fruitful discussions and words of advice. 
 This research used both the DeltaAI advanced computing and data resource, which is supported by the National Science Foundation (award OAC 2320345) and the State of Illinois, and the Delta advanced computing and data resource which is supported by the National Science Foundation (award OAC 2005572) and the State of Illinois \citep{delta}. Delta and DeltaAI are joint efforts of the University of Illinois Urbana-Champaign and its National Center for Supercomputing Applications. Additionally, this research used the Anvil supercomputer at Purdue University which is also supported by the National Science Foundation (award OAC 2005632) \citep{anvil}. Lastly, part of this work was funded by NSF fund CIF-2507523 and ARO fund 062982.

\end{ack}

\newpage
\bibliographystyle{plainnat}  
\bibliography{references}


\newpage
\appendix

\section{Hyperparameter Settings}
\label{app:appb}
\subsection{Computational Resources}
The experiments used NVIDIA GH200, H100, and L40S GPUs. All timing experiments were run on GH200s. The timing experiments also used 32 GB of RAM and 16 CPU cores. The CPU parallelism is useful for speeding up the Gurobi solver used during sparse Fourier optimization.

\subsection{Protein Design Methods}
\textbf{pre-trained}
We use the ProteinMPNN discrete diffusion model that was developed in the original DRAKES paper.

\textbf{DRAKES}
The KL-penalty is $\beta =0.001$. As with the pre-trained model, we use the RL fine-tuned ProteinMPNN model developed in the original DRAKES paper.

\textbf{Best-of-N}
The performance of Best-of-N will increase monotonically as $N$ increases, but with diminishing returns. We use Spectral Feedback in conjunction with Best-of-N to improve alignment beyond these limitations.

\subsection{Feedback Mechanisms}
Unless otherwise stated, we use $n=64$ samples for the value functions in Equation \ref{eq:f_tilde}.

\textbf{Feedback Loop}
We select $k=20$ edit-positions and $D=8192$ mask samples as our experimental parameters by following the results in the validation curves below. The alignment reward continues to improve with more edit-positions and more mask samples, though with diminishing returns. Increasing $k$ and $D$ also increases the computational cost so we found our selected parameters to be a reasonable balance. The mask sampling experiment does not use cross-validation during edit-selection, resulting in worse relative performance than the edit-positions experiment, which does use cross-validation. We used pre-selected settings (Max Depth=None, Leaves=50, Rate=0.01, $\lambda$=0.0001) because we faced compute resource limitations when increasing $D$ to 32768. We used 3 feedback iterations in these experiments for similar reasons.
\begin{figure}[H]
    \centering
    \includegraphics[width=0.4\linewidth]{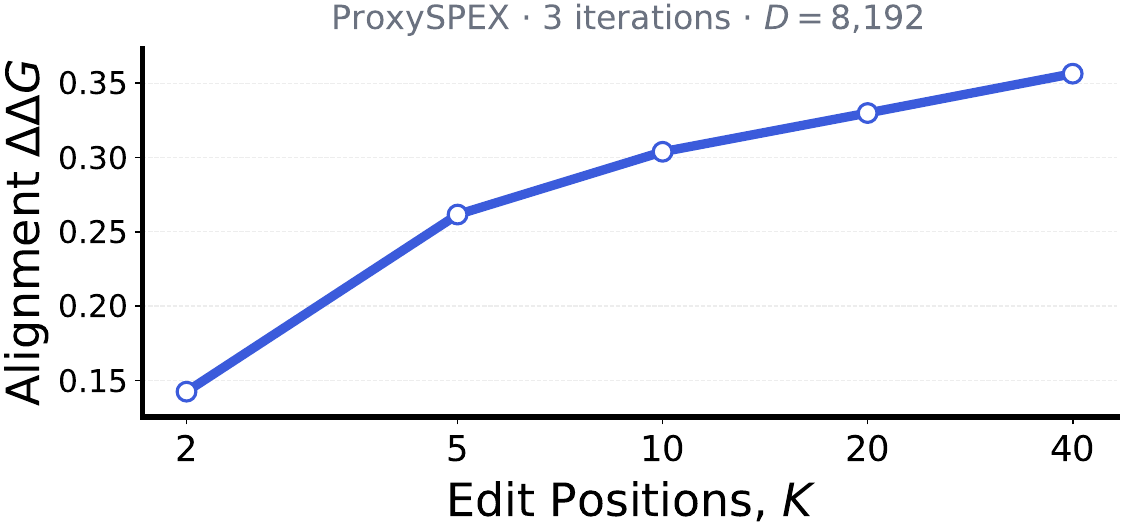}
    \hspace{2.5em}
    \includegraphics[width=0.4\linewidth]{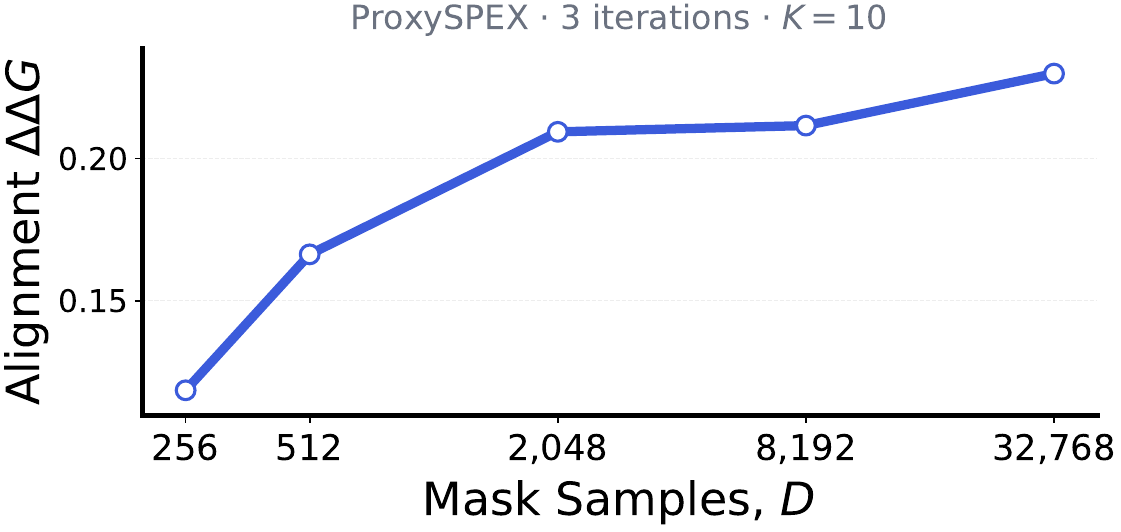}
    \caption{Validation curves for feedback loop parameters. Log-scale shows diminishing returns.}
    \label{fig:placeholder}
\end{figure}
\vspace{-1em}

\textbf{Spectral Feedback (ProxySPEX)}.
The ProxySPEX subroutine used in Spectral Feedback fits a GBT ensemble. We used the validation protein set, optimizing for each hyperparameter in isolation.
\begin{table}[H]
    \centering
    \begin{tabular}{|c|c|}
        \hline
        \textbf{Parameter} & \textbf{Validation Values}\\
        \hline
        Max. Tree Depth & $3, 5,    \text{None}$\\ \hline
        Number of leaves & $30, 50$\\
        \hline
        Learning Rate & $0.01, 0.1$ \\ \hline
        $\lambda$ & $5$ values with $\lambda_{min}=0.00001, \lambda_{max}=0.1, \lambda=0$ \\ \hline
    \end{tabular}
\end{table}

\textbf{First-order LASSO}. We use LASSO to implement a first-order sparse recovery method. In Spectral Feedback, we are given training sets $S_j \subseteq [L] \forall j \in [D]$. We define corresponding bit masks $b_j$ where $b_j[i] = 1 \Longleftrightarrow i \in S_j$. A first-order sparse recovery method models the value function as a linear function $f(S)=\beta + \sum_{i}c_i \cdot\mathbbm{1}\{i\in S\}$ where $\beta \in \mathbb{R}$ and $c\in \mathbb{R}^L$. We can then define the value function for the bit masks as $f(b) = \beta +  c^T b$. Each $S_j$ and $b_j$ have a corresponding training reward $r_j$. LASSO is natural for this setting because we assume a sparse Fourier transform and the $\ell_1$ norm induces sparsity. Applying LASSO with the penalty coefficient $\lambda \geq 0$ yields the following optimization problem:
\[
(\beta, \tilde{c}) = \underset{\beta\in\mathbb{R},\;c\in \mathbb{R}^L}{\arg \min}\; \frac{1}{D}\sum_{j=1}^D(\beta + c^Tb_j-r_j)^2 + \lambda ||c||_1
\]

We can then write the maximization problem of $f(b)$ like so:
\[
b^* = \underset{b\in \{0,1\}^L,\;\sum _i b[i] \leq k}{\arg\max} \tilde{c}^Tb
\]
The solution is then the top-$k$ coefficients of $\tilde{c}$ that are also positive:
\[
b^*[i] = \left\{
\begin{aligned}
&1 \text{ if } \tilde{c}_i > 0 \text{ and } \tilde{c}_i \in \text{top $k$ coefficients}\\ 
&0 \text{ else }  
\end{aligned}
\right.
\]

As with Spectral Feedback, when using LASSO for edit-position selection, we run cross-validation to be consistent with the original ProxySPEX paper \citep{butler2025proxyspexinferenceefficientinterpretabilitysparse}. 
\begin{table}[H]
\centering
\begin{tabular}{|c|c|}
    \hline
    \textbf{Parameter} & \textbf{Validation Values}\\ \hline
    $\lambda$ & $5$ values with $\lambda_{min}=0.00001, \lambda_{max}=0.1$ and $\lambda=0.0$\\ \hline
\end{tabular}
\end{table}

\textbf{Exclusion \& Gradient}. The only hyperparameter for these edit-position methods is the number of edit-positions, which we already select for Spectral Feedback on the validation set. We use the same number of positions (20 in our experiments) to make them fair baselines.

\section{Discrete 
Diffusion Models}
\label{app:appendixa}
The discrete diffusion model is a type of generative language model that has been popularized for biology-related tasks such as inverse folding. We describe these models formally and explain how to sample from them. First, consider a finite vocabulary $V$, a sequence length $L$, and a corresponding sample space $\mathcal{X}:=V^L$. We then model a corresponding probability mass trajectory $p_t\in\mathbb{R}^N$ where $N:=|\mathcal{X}|=|V|^L$. Like in continuous diffusion, the discrete diffusion process involves evolving $p_t$ by a differential equation. The evolution of $p_t$ forward in time is called the forward process. In the continuous case, it is common to use the Fokker-Planck equation, a partial differential equation related to the Itô diffusion process. In the discrete setting, the following linear ordinary differential equation is used:
\begin{align*}
    \frac{dp_t}{dt}=Q_tp_t, \quad p_0 = p_{data}
\end{align*}
Here, $Q_t\in \mathbb{R}^{N\times N}$ are the rate matrices of a Continuous Time Markov Chain (CTMC) and correspondingly must satisfy
\begin{align*}
    Q_t(i,j)&\geq 0 \quad \forall i \neq j\\
    \sum_i Q_t(i,j)&=0 \quad \forall j.
\end{align*}
The first constraint is due to the non-negativity of probability masses, whereas the second constraint ensures that the total probability mass of the system does not change. Like for continuous diffusion models, to make such a process useful for inference, we need a way to reverse the process. In particular, if we structure the process to converge to a probability distribution $p_T$ from which we know how to sample, and then sample from the reverse process initialized by $p_T$, we will converge to the target distribution $p_0$. The following is a commonly used reverse process \citep{liang2025discretediffusionmodelsnovel}:
\begin{align*}
    \frac{dp_{T-t}}{dt}=\bar{Q}_{T-t}p_{T-t}, \quad \bar{Q}_t(x,y)=\left\{
    \begin{aligned}
        \frac{p_t(x)}{p_t(y)}Q_t(y,x)&\quad x \neq y\\
         -\sum_{x'\neq x}\bar{Q}_t(x',x) &\quad x=y
    \end{aligned}\right.
\end{align*}
Like before, the reverse process is a CTMC and $\bar{Q}_t$ are the corresponding rate matrices. We can design the process to converge to a distribution of our choosing by selecting $Q_t$ accordingly. Thus, to learn $\bar{Q}_t$ so that we can sample from the reverse process, we just need to learn $p_t(y)/p_t(x)$. This ratio is commonly called the score function \citep{sun2023scorebasedcontinuoustimediscretediffusion}.

In its most abstract form, designing a discrete diffusion model requires defining a forward process with some generator $Q_t$ and then learning a corresponding score function $p_t(y)/p_t(x)$ from the training data. A popular choice for the forward process is random masking. The CTMC has some probability of transitioning tokens into masks, and once a token is masked, it stays masked. The reverse process is initialized with a fully masked sequence $x_T$, and then progressively flips tokens from masks to unmasked elements in the vocabulary. Analogous to the forward process, a token will stay unmasked once it leaves its masked state.

The Euler-Maruyama scheme, a first-order accurate iterative method, is commonly used to approximate the solution of continuous diffusion models. It is then natural to apply Euler's method to approximate the reverse discrete diffusion evolution:
\begin{align*}
    p_{n+1}\leftarrow p_{n}+h\bar{Q}_{T-t_{n}}p_n
\end{align*}
where $h$ is the step size. This method is unfortunately impractical since it would require keeping track of the scores for an exponential number of state transitions $x\rightarrow y$. Instead, the following Hamming distance constraint is often used: $Q_t(x,y)=0$ if $d(x,y)>1$ \citep{liang2025discretediffusionmodelsnovel}. This allows only state transitions that flip one token. Still, to support multiple token updates during a single step for faster inference, the transitions of each token are assumed to be independent.

\section{Compute Complexity}
\label{app:compute_complexity}
\begin{figure}[H]
    \centering
    \includegraphics[width=0.9\linewidth]{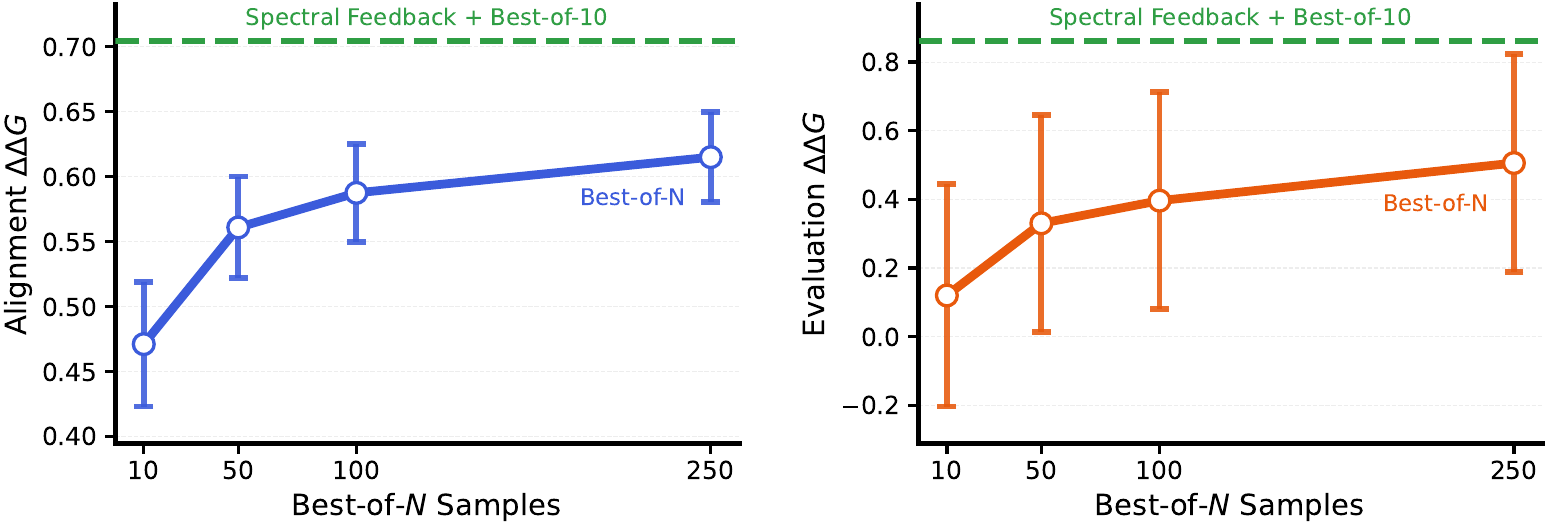}
    \caption{Spectral Feedback with Best-of-10 reaches better performance than scaling Best-of-N. Hyperparameters: $D=8192$, $k=20$, and 5 feedback iterations.}
    \label{fig:bon_diminishing}
\end{figure}

The Spectral Feedback pipeline involves calls to several models with varying computational costs. The protein diffusion model is called at each iteration of the reverse process. In the first iteration, a protein sequence is generated from a fully masked sequence. Subsequent iterations initialize the reverse process with partially noisy sequences via the feedback process. Recall that $N$ is the total number of generated proteins so the number of feedback iterations is $N - 1$. In each feedback iteration Spectral Feedback makes $D$ calls to the mask value function. Let $M$ be the number of reverse process steps. Then Spectral Feedback makes $M \times N$ calls to the diffusion model and $D \times (N - 1)$ calls to the mask value function. The value functions in Equation \ref{eq:f_tilde} call the diffusion model once to get state change probabilities of a single reverse process step, then sample $n$ state changes and pass each generated sequence to the alignment reward oracle. With these value functions, Spectral Feedback makes $M\times N + D\times (N-1)$ diffusion model calls and $n \times D \times (N - 1)$ alignment reward oracle calls. Best-of-N is cheaper in compute since it makes $M\times N$ calls to the diffusion model and $N$ calls to the alignment reward oracle. A key limitation of Best-of-N, however, is that it has diminishing returns as $N$ increases and levels off in reward. With protein design, desirable proteins may be rare as discussed in 
\citet{proteinguide}. Figure~\ref{fig:bon_diminishing} demonstrates that Spectral Feedback can improve Best-of-N beyond its diminishing returns and identify these rare yet more stable proteins. The modularity of the algorithm allows it to also improve other alignment methods such as Beam Search and RL as shown in Figure~\ref{fig:spec_feedback_applications}.

\newpage
\section{Sparse Fourier Approximation}
\subsection{Sparse Fourier Optimization}
\label{supp:optimization}
The derivation in this section originally appeared in Appendix A.3 of \citet{butler2025proxyspexinferenceefficientinterpretabilitysparse}. We restate it here.

Assume $\hat{f}: 2^{[n]} \to \mathbb{R}$ has a sparse, low-degree Fourier expansion with support
\[
    \mathcal{A} \;=\; \bigl\{\,T \subseteq [n] \,:\, \hat{F}(T) \neq 0,\; |T| \le d\,\bigr\},
    \qquad |\mathcal{A}| \ll 2^n,
\]
such that
\[
    \hat{f}(S) \;=\; \sum_{T \in \mathcal{A}} (-1)^{|S \cap T|}\,\hat{F}(T).
\]
To formulate the optimization as an integer program, we re-express $\hat{f}$ in the M\"obius basis, which replaces the parity functions $(-1)^{|S \cap T|} \in \{-1, +1\}$ with subset-indicator functions $\mathbbm{1}[T \subseteq S] \in \{0, 1\}$. Through a change of variable, the M\"obius coefficients are
\[
    \hat{M}(T) \;=\; (-2)^{|T|} \sum_{\substack{S \in \mathcal{A}, \\ S \supseteq T}} \hat{F}(S).
\]
Letting $\mathcal{A}^{+} = \bigl\{\, R \subseteq T \,\bigm|\, T \in \mathcal{A} \,\bigr\}$ denote the downward closure of $\mathcal{A}$, the inverse M\"obius transform gives
\[
    \hat{f}(S) \;=\; \sum_{\substack{R \in \mathcal{A}^{+}, \\ R \subseteq S}} \hat{M}(R).
\]
The optimization problem can then be expressed as a polynomial over $\{0,1\}^n$. Let $\mathbf{x} \in \{0,1\}^n$ be the characteristic vector of $S$, so that $x_i = 1$ if and only if $i \in S$. We focus on the maximization problem (minimization follows analogously):
\[
    \max_{\substack{S \subseteq [n], \\ |S| \le k}} \hat{f}(S)
    \;=\;
    \max_{\substack{\mathbf{x} \in \{0,1\}^n, \\ \sum_i x_i \le k}}
    \sum_{R \in \mathcal{A}^{+}} \hat{M}(R) \prod_{i \in R} x_i.
    \label{eq:poly-opt}
\]

To reduce the problem to a linear integer program, each monomial $\prod_{i \in R} x_i$ is replaced with a binary decision variable $y_R \in \{0,1\}$. We augment $\mathcal{A}^{+}$ to include all singletons $\{i\}$ for $i \in [n]$, setting $\hat{M}(\{i\}) = 0$ for any newly added singletons, so that $y_{\{i\}}$ exists for every $i$. Linking constraints enforce $y_R = \prod_{i \in R} x_i$:
\begin{alignat*}{2}
    \max_{\mathbf{y} \in \{0,1\}^{|\mathcal{A}^{+}|}}
    \quad & \sum_{R \in \mathcal{A}^{+}} \hat{M}(R)\, y_R \\[4pt]
    \text{s.t.}\quad
    & y_R \;\le\; y_Q
        &&\quad \forall\, Q \subset R,\; R \in \mathcal{A}^{+} \\[2pt]
    & \sum_{i \in R} y_{\{i\}} \;\le\; |R| - 1 + y_R
        &&\quad \forall\, R \in \mathcal{A}^{+} \\[2pt]
    & \sum_{i \in [n]} y_{\{i\}} \;\le\; k.
\end{alignat*} The first constraint guarantees that whenever a monomial is activated (i.e., $x_i = 1$ for all $i \in R$), all of its subsets are also activated. The second ensures that if a monomial is deactivated (i.e., $x_i = 0$ for some $i \in R$), then at least one of its constituent singletons $y_{\{i\}}$ is likewise deactivated. The third imposes the cardinality constraint, and after solving the program, the solution $\mathbf{x}$ is read off from $y_{\{i\}}$.

Let $s = |\mathcal{A}|$ be the sparsity of $\hat{f}$. The resulting program has at most $s \cdot 2^{d} + n$ decision variables: up to $2^{d}$ subsets per element of $\mathcal{A}$, plus the $n$ augmented singletons. The constraints decompose as at most $(s \cdot 2^{d} + n)(2^{d} - 1)$ subset-activation constraints (one per pair $Q \subset R$), $s \cdot 2^{d} + n$ deactivation constraints, and a single cardinality constraint, giving at most $s \cdot 4^{d} + n \cdot 2^{d} + 1$ constraints in total. The program is therefore tractable when $\hat{f}$ is both sparse ($s$ small) and
low-degree ($d$ small), even for large $n$.

We solve the program using Gurobi's default branch-and-cut algorithm via the \texttt{gurobipy} interface \citep{gurobi}.

\newpage
\subsection{Sparsity Experiments}
\label{sec:r2_exps}
\raggedbottom

The data in Figure~\ref{fig:sparsity} were computed with SPEX under a compute budget of 100k and a max interaction order of 5. The value functions use $n=64$ empirical samples for each evaluation. The underlying protein model is the pre-trained model. We show additional spectral analysis results below for each protein in the test set, using both the $f_{max}$ and the $f_{avg}$ value functions. We evaluate the sparsity of the functions as well as the necessity of high-order terms. We first study the $\Delta\Delta G$ alignment oracle that was used in the paper's main results.
We repeat this study with the ProtGPT2 Log-Likelihood oracle as the target alignment oracle. The results show that both value functions exhibit sparse Fourier transforms as they can be approximated with high faithfulness ($R^2$) by using relatively few Fourier coefficients. Additionally, most spectral variance is concentrated at lower interaction orders. Interestingly, $f_{avg}$ tends to achieve a higher $R^2$ than $f_{max}$ under the same compute budget. This may reflect greater variance in $f_{max}$ than in $f_{avg}$.

\begingroup
\center{\textsc{$\Delta\Delta G$ Alignment Oracle}}
\vspace{1em}

\setlength{\intextsep}{3pt}

\newcommand{\ddgrow}[3]{%
\begin{figure}[H]
  \centering
  \includegraphics[width=0.32\linewidth]{Figures/neurips_rebuttal_figs/exps#2_#3/exps#2_#1_sparsity_#3.pdf}%
  \hfill
  \includegraphics[width=0.32\linewidth]{Figures/neurips_rebuttal_figs/exps#2_#3/exps#2_#1_maxorder_#3.pdf}%
  \hfill
  \includegraphics[width=0.32\linewidth]{Figures/neurips_rebuttal_figs/exps#2_#3/exps#2_#1_variance_#3.pdf}%
\end{figure}
\vspace{-0.5\baselineskip}
}

\newcommand{\ddgsmall}[3]{%
\begin{minipage}{0.49\linewidth}
  \centering
  \includegraphics[width=0.32\linewidth]{Figures/neurips_rebuttal_figs/exps#2_#3/exps#2_#1_sparsity_#3.pdf}%
  \hfill
  \includegraphics[width=0.32\linewidth]{Figures/neurips_rebuttal_figs/exps#2_#3/exps#2_#1_maxorder_#3.pdf}%
  \hfill
  \includegraphics[width=0.32\linewidth]{Figures/neurips_rebuttal_figs/exps#2_#3/exps#2_#1_variance_#3.pdf}%
\end{minipage}%
}

\ddgrow{r6_560_TrROS_Hall}{3}{ddg}

\ddgrow{v2K43S_2KVV}{3}{ddg}

\begin{figure}[H]
  \centering
  \ddgsmall{1F0M}{3}{ddg}\hfill\ddgsmall{2L09}{3}{ddg}\\[3pt]
  \ddgsmall{2LVN}{3}{ddg}\hfill\ddgsmall{2M2J}{3}{ddg}\\[3pt]
  \ddgsmall{2MA4}{3}{ddg}\hfill\ddgsmall{4G3O}{3}{ddg}\\[3pt]
  \ddgsmall{5JRT}{3}{ddg}\hfill\ddgsmall{7JJK}{3}{ddg}\\[3pt]
  \ddgsmall{HEEH_KT_rd6_0746}{3}{ddg}\hfill\ddgsmall{2KRU}{3}{ddg}
\end{figure}

\begin{center}
\textsc{ProtGPT2 Log-Likelihood Oracle}
\end{center}
\vspace{1em}

\ddgrow{r6_560_TrROS_Hall}{6}{ll}

\ddgrow{v2K43S_2KVV}{6}{ll}

\begin{figure}[H]
  \centering
  \ddgsmall{1F0M}{6}{ll}\hfill\ddgsmall{2L09}{6}{ll}\\[3pt]
  \ddgsmall{2LVN}{6}{ll}\hfill\ddgsmall{2M2J}{6}{ll}\\[3pt]
  \ddgsmall{2MA4}{6}{ll}\hfill\ddgsmall{4G3O}{6}{ll}\\[3pt]
  \ddgsmall{5JRT}{6}{ll}\hfill\ddgsmall{7JJK}{6}{ll}\\[3pt]
  \ddgsmall{HEEH_KT_rd6_0746}{6}{ll}\hfill\ddgsmall{2KRU}{6}{ll}
\end{figure}

\endgroup

\newpage
\subsection{High-Sensitivity Value Function Ablation}
\label{app:exp_tilt_val_func}

High global \(R^2\) is not by itself a sufficient certificate for argmax recovery. One can easily construct counterexamples where an approximation has arbitrarily high global \(R^2\), yet selects a poor maximizer. Given our strong experimental results using the global-\(R^2\)-driven approximation, we do not believe our value functions fall into this pathological regime. Nevertheless, this is an important distinction. Alternative objectives can be used to more directly align with maximization. In particular, a value function \(f(S)\) can be replaced with a strictly increasing transformation
\[
g(S) = \phi(f(S)),
\]
which preserves the optimizer but can be used to make an \(L_2\) approximation increasingly sensitive to high-value subsets. One concrete choice is an exponentially tilted objective,
\[
g_\beta(S)
=
\exp\left(
\beta \frac{f(S)-f_{\max}}{f_{\max}-f_{\min}}
\right),
\]
where \(f_{\max}\) and \(f_{\min}\) denote the maximum and minimum values of
\(f\) over the sampled sets. Under this normalization, the maximizer has value
\(1\), while lower-value subsets are exponentially downweighted toward
\(\exp(-\beta)\). Larger \(\beta\) therefore places more emphasis on accurately
approximating the top of the value function. We use the convention that for $\beta=0$, $g_0(S)=f(S)$.

This transformation has interesting implications for spectral sparsity. For normalization with unique global extrema, as
\(\beta \to \infty\), \(g_\beta\) approaches a Dirac delta at the maximizer
\(S^\star\). A Dirac delta on the Boolean cube is spectrally dense. We have repeated our experiments above by analyzing spectral sparsity under different severities of exponential tilt. We find that for modest choices of \(\beta\) (0.25-8), the sparsity levels of the function remain similar to the untransformed \(f(S)\), and the proportion of spectral variance in higher-degree terms remains high. For larger values of \(\beta\), the spectrum places more energy on high-order terms and becomes increasingly dense, making it difficult for spectral recovery techniques to achieve high \(R^2\). Thus, increasing \(\beta\) can trade better argmax alignment for weaker Fourier sparsity. An improved version of Spectral Feedback could try learning \(g_\beta\) for a range of $\beta$ using the same samples, and then cross-validate to select the highest \(\beta\) for a pre-specified approximation quality.

\begin{figure}[H]
    \centering
    \includegraphics[width=0.9\linewidth]{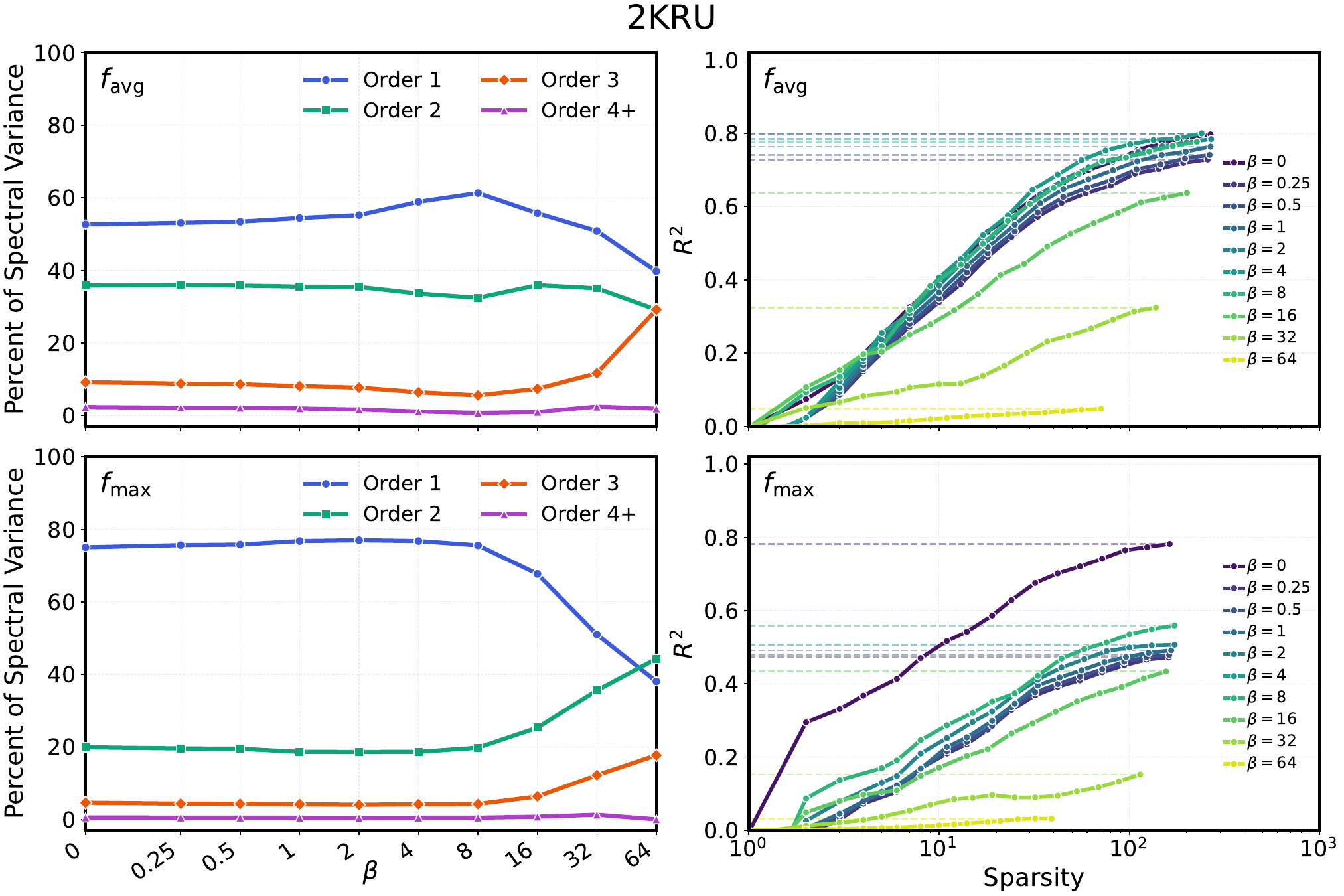}
    \caption{Exponential tilt function Fourier analysis for 2KRU with the pre-trained model.}
    \label{fig:2kru_beta_tilt}
\end{figure}
\vspace{-1em}
\begin{figure}[H]
    \centering
    \includegraphics[width=0.9\linewidth]{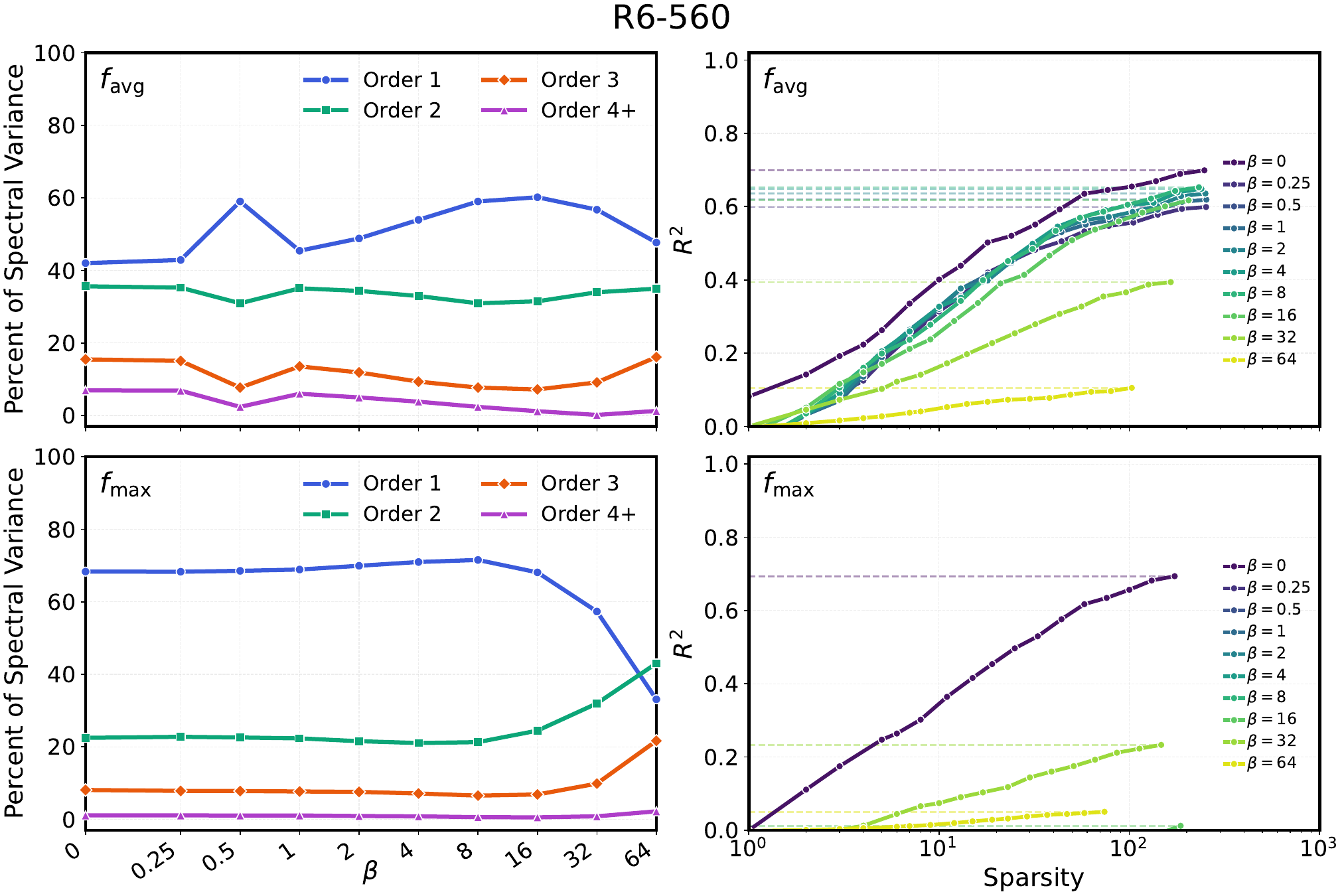}
    \caption{Exponential tilt function Fourier analysis for r6\_560\_TrROS\_Hall with pre-trained model.}
    \label{fig:r6_beta_tilt}
\end{figure}

\section{Spectral Feedback $scRMSD$ Discussion}
\label{sec:scrmsd_drakes_disc}
Table~\ref{fig:spec_feedback_modular_table} shows an increase in $scRMSD$ when Spectral Feedback is applied to DRAKES. This is an undesired result since this means the inverse folding is less accurate. Recall that one of the two conditions for a protein to be classified as successful is that $scRMSD <2$. An increase in $scRMSD$ then limits the gains that can be made in protein success rate, as seen in Figure~\ref{fig:spec_feedback_modular_plots}. 

To better understand this behavior, we analyzed scatter and contour plots relating $scRMSD$ to alignment $\Delta \Delta G$.
The results in the corresponding figure below indicate that Spectral Feedback inherits the properties of both the underlying model and the reward objective. In the DRAKES distribution, the highest-reward region contains structurally inconsistent samples, including sequences with $scRMSD$ values between 8 and 10, well above the desired threshold of 2. In contrast, the pre-trained model has fewer of these high-$scRMSD$ samples, and its highest-reward samples generally retain lower $scRMSD$. As a result, the average $scRMSD$ is greater for DRAKES than for the pre-trained model: 1.17 vs 1.10. This difference increases with the addition of Spectral Feedback as reported in the paper. DRAKES also has a lower average log-likelihood than the pre-trained model, as seen in Table~\ref{fig:spec_feedback_modular_table}, which further indicates overfitting to the alignment oracle that leads to more out-of-distribution proteins.

\begin{figure}[H]
    \centering
    \includegraphics[width=0.48\linewidth]{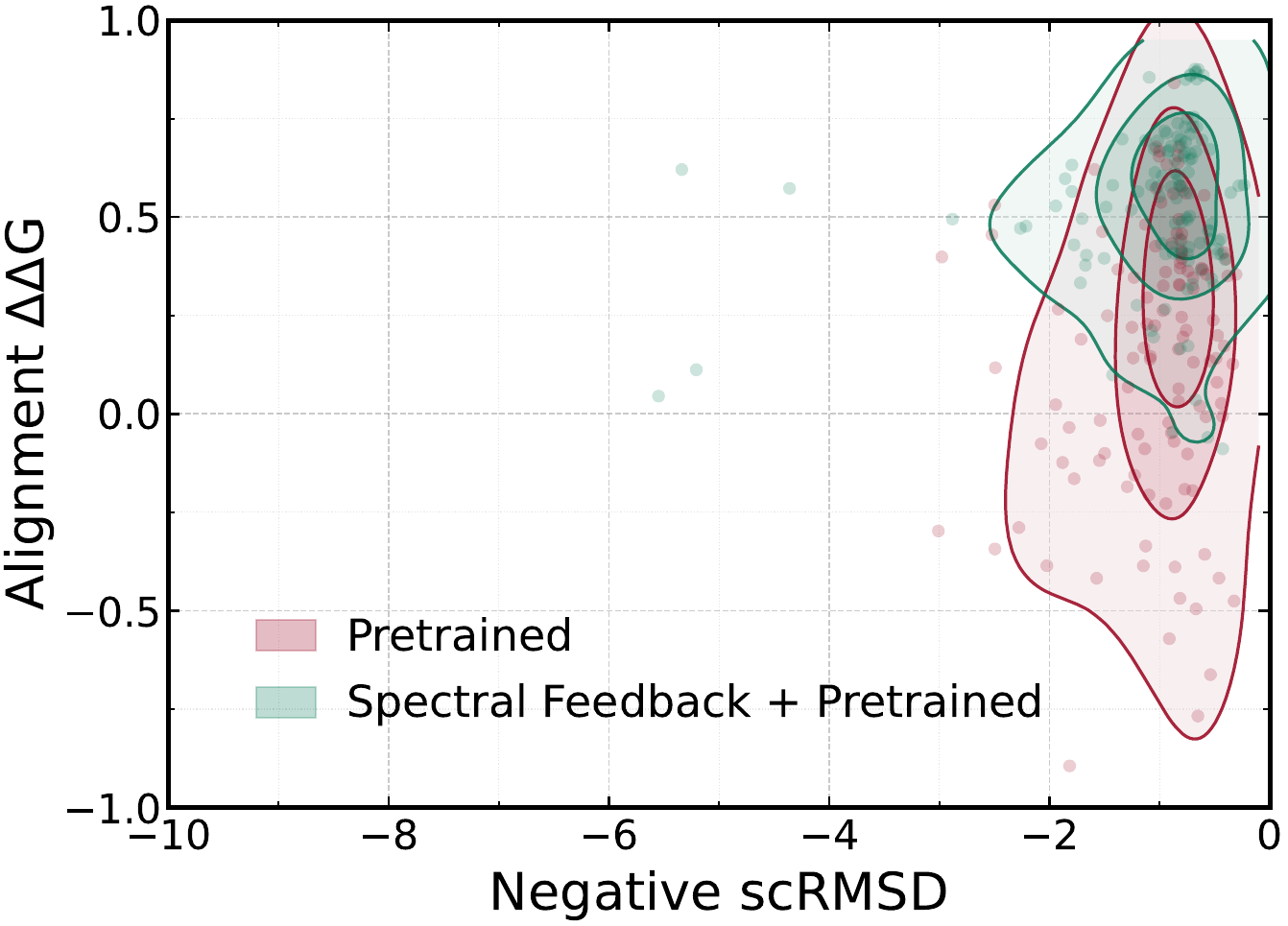}
    \includegraphics[width=0.48\linewidth]{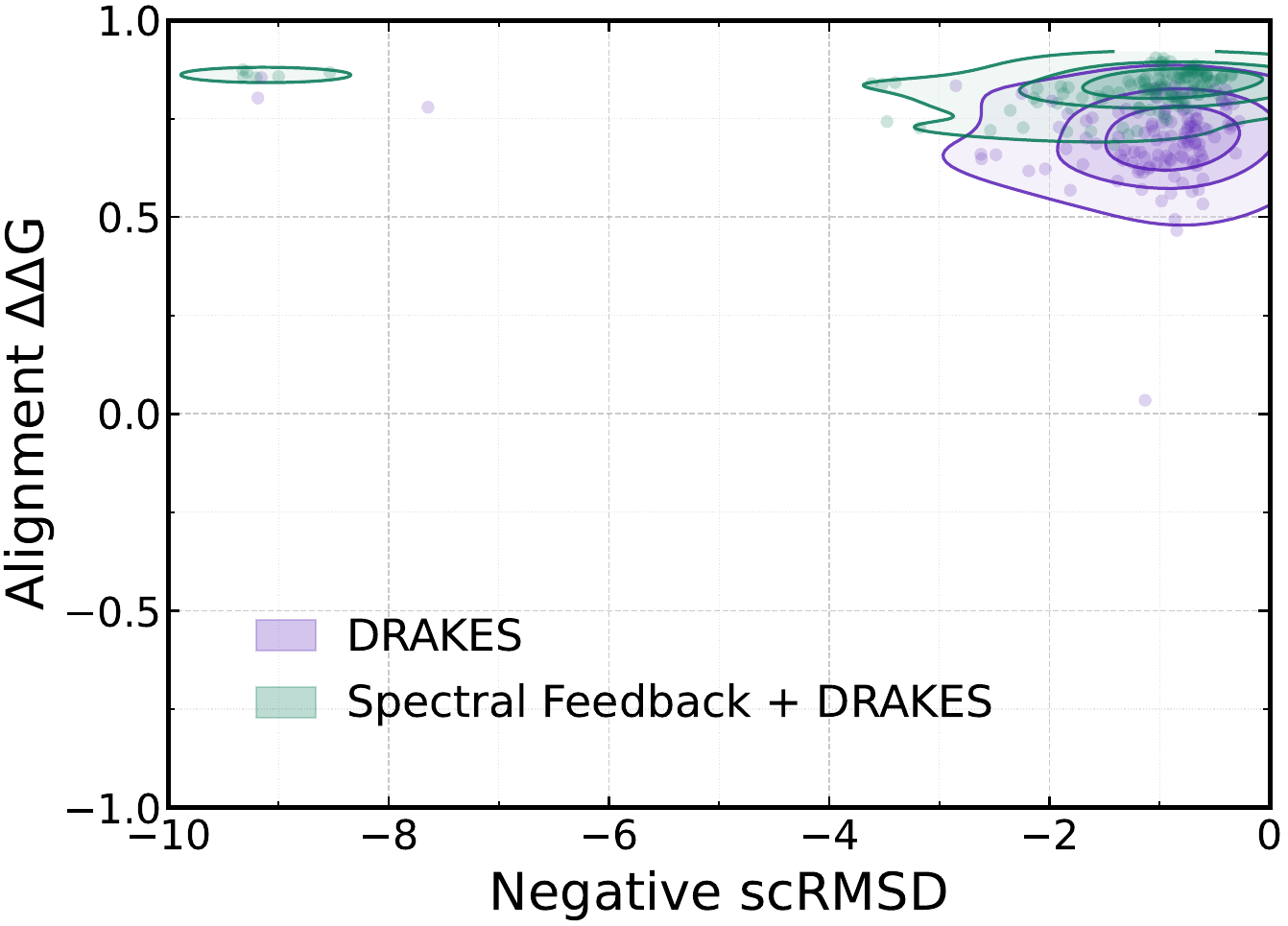}
    \caption{$scRMSD$ vs $\Delta \Delta G$ distributions show a worse trade-off with $scRMSD$ for DRAKES when applying Spectral Feedback.}
    \label{fig:contour_plot}
\end{figure}

Spectral Feedback further optimizes the stability reward and therefore can move samples toward regions where stability reward improves but structural consistency degrades. However, this degradation is likely a reflection of the DRAKES distribution in high-reward regions. We conclude that Spectral Feedback improves a specified alignment objective, while preservation of auxiliary properties depends on the quality of the reward and the distribution induced by the underlying model. In settings such as DRAKES, a regularized or multi-objective reward that explicitly includes structural consistency or correlated statistics would be needed to prevent this trade-off. We discuss using a multi-objective reward in Appendix~\ref{app:additional_alignment_oracles}.

\section{Additional Scaling and Alignment Studies}
\subsection{Spectral Feedback Applications}

\begin{figure}[H]
    \centering
    \includegraphics[width=0.65\linewidth]{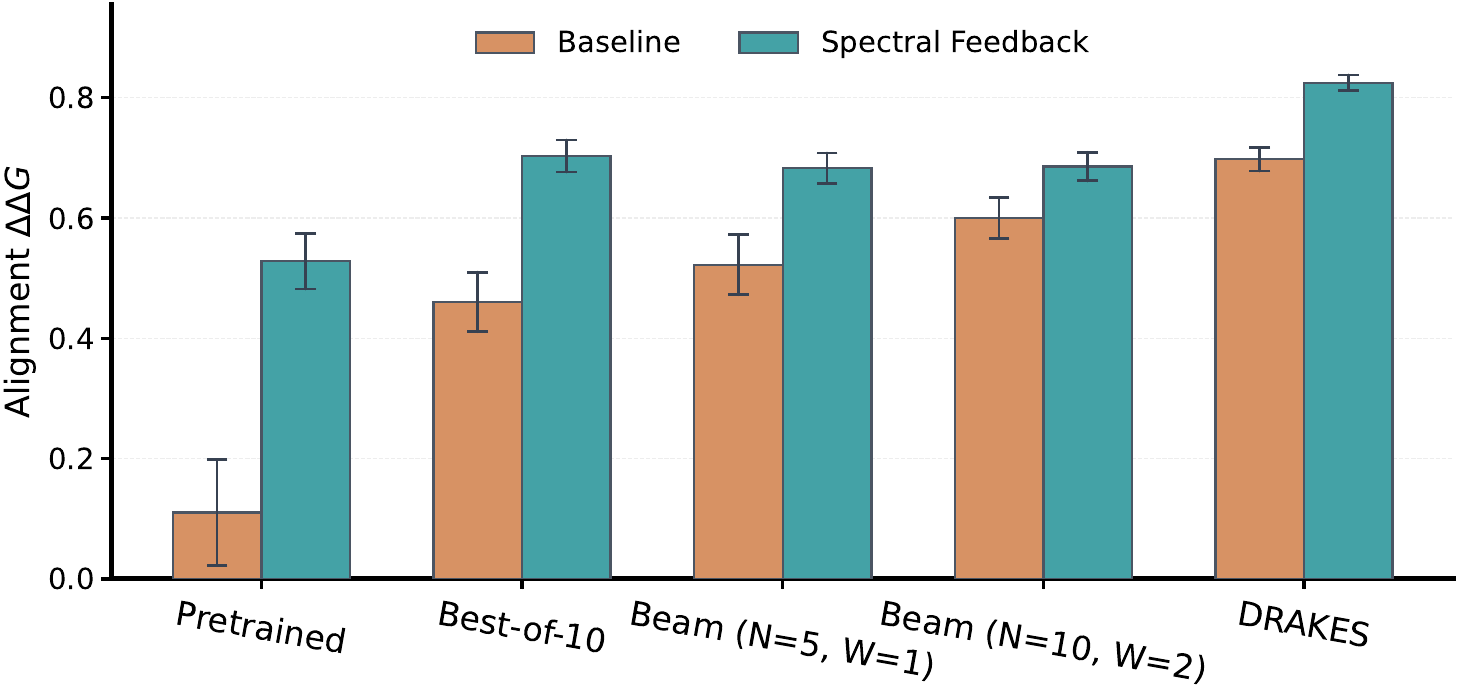}
    \caption{Spectral Feedback with $k=20$, $D=8192$, and 5 feedback iterations.}
    \label{fig:spec_feedback_applications}
\end{figure}

In general, feedback can use substantial compute resources when many samples are used to make edit-selections; however, it is useful when a target alignment model reaches insufficient reward values. Still, it is important to make each feedback step as efficient as possible and Figure~\ref{fig:baseline_edit_position_selection_curve} shows that Spectral Feedback with ProxySPEX scales well in alignment performance and latency when compared to competing edit-position selection methods.

\subsection{Additional Alignment Oracles}
\label{app:additional_alignment_oracles}

Most of this paper focuses on aligning inverse protein folding diffusion models with a $\Delta\Delta G$ reward oracle, though it is important to study the flexibility of Spectral Feedback across not just different protein models but also different reward oracles. We study its performance with two additional alignment oracles: ProtGPT2 Log-Likelihood and a multi-objective reward that balances log-likelihood with $\Delta \Delta G$.

\begin{center}
    \textbf{ProtGPT2 Log-Likelihood Oracle}
\end{center}

The ProtGPT2 Log-Likelihood oracle computes log-likelihoods of diffusion model sample sequences by using the ProtGPT2 autoregressive language model. This has been done in previous works such as \citet{gruver2023proteindesignguideddiscrete} and \citet{cemri2024discrete}. These log-likelihoods can be used as a measure of naturalness of protein sequences because ProtGPT2 was trained to capture the distribution of protein amino acid sequences \citep{ferruz_protgpt2_2022}. The figure below shows Spectral Feedback applied to the pre-trained model with the ProtGPT2 oracle, using first-order LASSO as the edit-selection strategy.

\begin{figure}[H]
    \centering
    \includegraphics[width=0.5\linewidth]{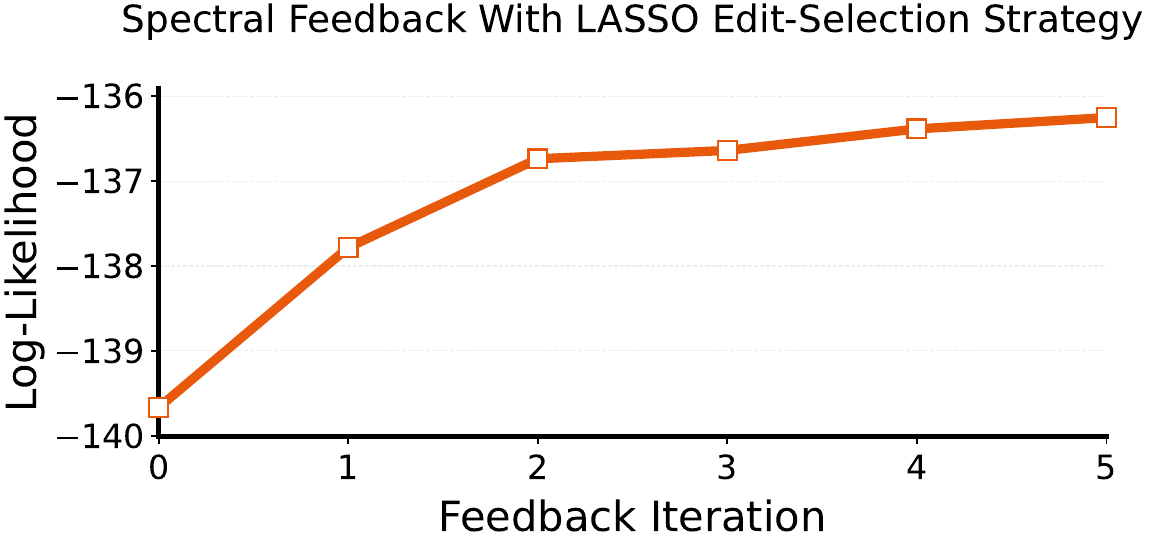}
    \caption{ProtGPT2 Log-Likelihood alignment trajectory with Spectral Feedback. Hyperparameters are $k=10$, $D=2048$, $n=16$ (reward oracle calls per value function call).}
    \label{fig:placeholder4}
\end{figure}

\begin{center}
\textbf{Balanced Oracle}
\end{center}

We define the multi-objective reward oracle as $r(x)=\alpha \cdot \Delta\Delta G(x)+0.005\cdot(1-\alpha)\cdot ProtGPT2(x)$. The weight $\alpha$ balances between aligning for $\Delta \Delta G$ and aligning for log-likelihood. The constant $0.005$ rescales the ProtGPT2 oracle to make the choice of $\alpha$ more linear for weighting the importance of the two oracles. We show results for two proteins in the test set which demonstrate differing trade-offs. R6-560 has a direct trade-off between the oracles whereas 2KRU has higher log-likelihoods when primarily aligning for $\Delta \Delta G$.

\begin{figure}[H]
    \centering
    \includegraphics[width=\linewidth]{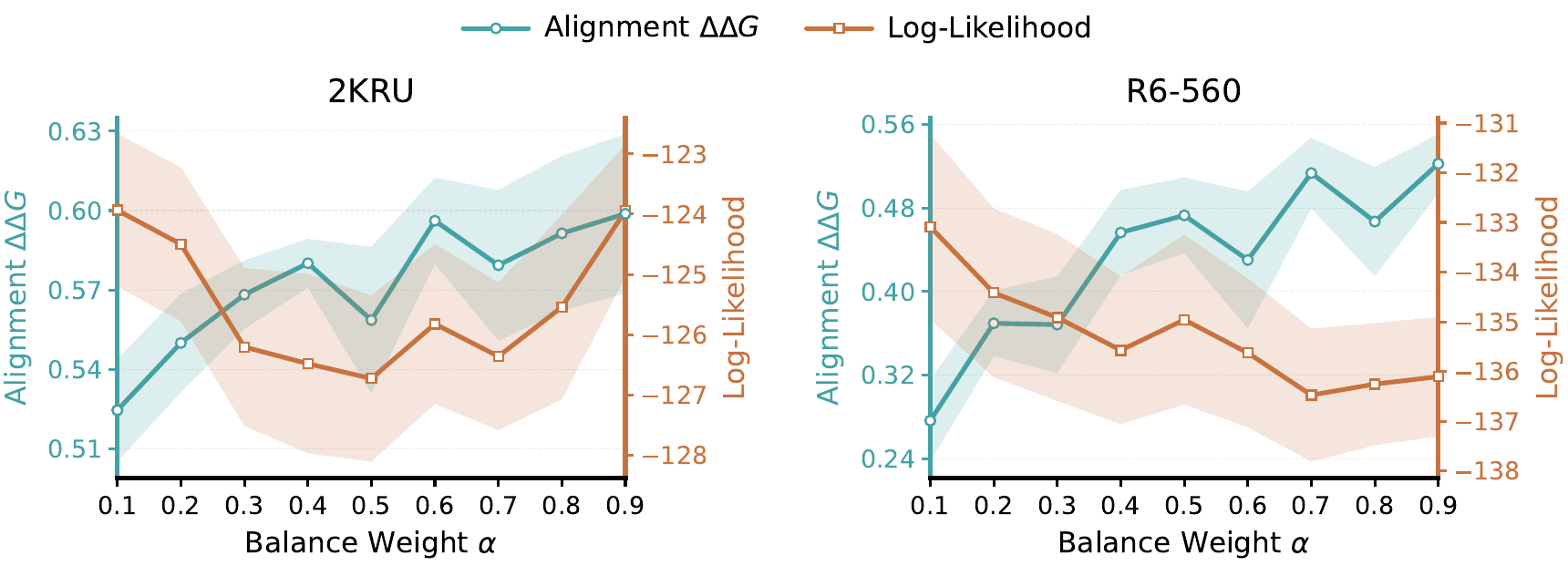}
    \caption{An example multi-objective oracle balances between $\Delta \Delta G$ and ProtGPT2 Log-Likelihood.}
    \label{fig:placeholder5}
\end{figure}

These results show that a multi-objective oracle can be useful for targeting and balancing between multiple objectives with Spectral Feedback. This can be useful as a form of regularization to avoid overfitting, which is typical for alignment methods with a maximizing objective. For example, one could make an oracle that balances between $\Delta \Delta G$ and $scRMSD$ to avoid the high $scRMSD$ proteins identified in Figure~\ref{fig:contour_plot}.

\section{High-Order Interactions Case Study ($\Delta \Delta G$ Alignment)}
\label{app:spec_study}
\subsection{Comparing ProxySPEX with LASSO}
In our work, we observed that the necessity of multivariate interactions for determining edit-positions is highly dependent on the target protein. While Spectral Feedback performs well when paired with ProxySPEX, a high-order sparse recovery method such as ProxySPEX may not always be necessary. We study how LASSO performs as a first-order sparse recovery method to better understand the impact of high-order sparse recovery. We first compare using ProxySPEX and LASSO for Spectral Feedback with two protein backbones: R6-560 and 2KVV. We run a single Spectral Feedback process for each backbone and initialize each process with the same corresponding protein sequences that were studied in Figure~\ref{fig:sparsity}. We then compare ProxySPEX with LASSO across the entire test set.

In the first experiment, ProxySPEX had an 8.3\% larger final reward than LASSO for \textit{2KVV} and a 24\% larger reward for \textit{R6-560}. A possible explanation for these results lies in the structural differences between the protein backbones.
Figure~\ref{fig:spec_feedback_modular_plots} shows that the \textit{2KVV} backbone contains notable $\alpha$-helices while this structure is less prominent in \textit{R6-560}. An $\alpha$-helix is a secondary protein structure that is formed by structured local interactions between the $i^{th}$ and $(i+4)^{th}$ amino acids in a protein sequence \citep{alpha_helix_chapter}. The primarily local interactions that create this structure may then lead to less complex interactions between edit-positions. In Figure~\ref{fig:sparsity}, for $f_{avg}$, \textit{2KVV} has strong first-order interactions while \textit{R6-560} has more significant high-order interactions that ProxySPEX can leverage. Differences in backbone structure may influence differences in the value function Fourier spectra, which then relate to the performance of ProxySPEX and LASSO for edit-selection. We show detailed results of the edit-selections and amino-acid updates for this experiment in Appendices \ref{app:protein_case_study_exp_r6_560} and \ref{app:protein_case_study_exp_2kvv}.

In the second experiment, ProxySPEX slightly outperforms LASSO for the pre-trained model and Best-of-10. Additionally, LASSO slightly outperforms ProxySPEX for DRAKES. The close performance may be because most proteins in the test set have dominant first-order interactions for both value functions (see Appendix~\ref{sec:r2_exps}). It would be interesting to see if the corresponding spectral profiles of the value functions for DRAKES are also more dominantly first-order, given the better performance of LASSO. It would also be interesting to find cases where high-order interactions are more dominant and where ProxySPEX would stand out more. Perhaps better performance could be reached in cases where the value function is deterministic since then larger $R^2$ values would be achievable and the sparse recovery from ProxySPEX would be more accurate.

\begin{figure}[H]
    \centering
    \includegraphics[width=\linewidth]{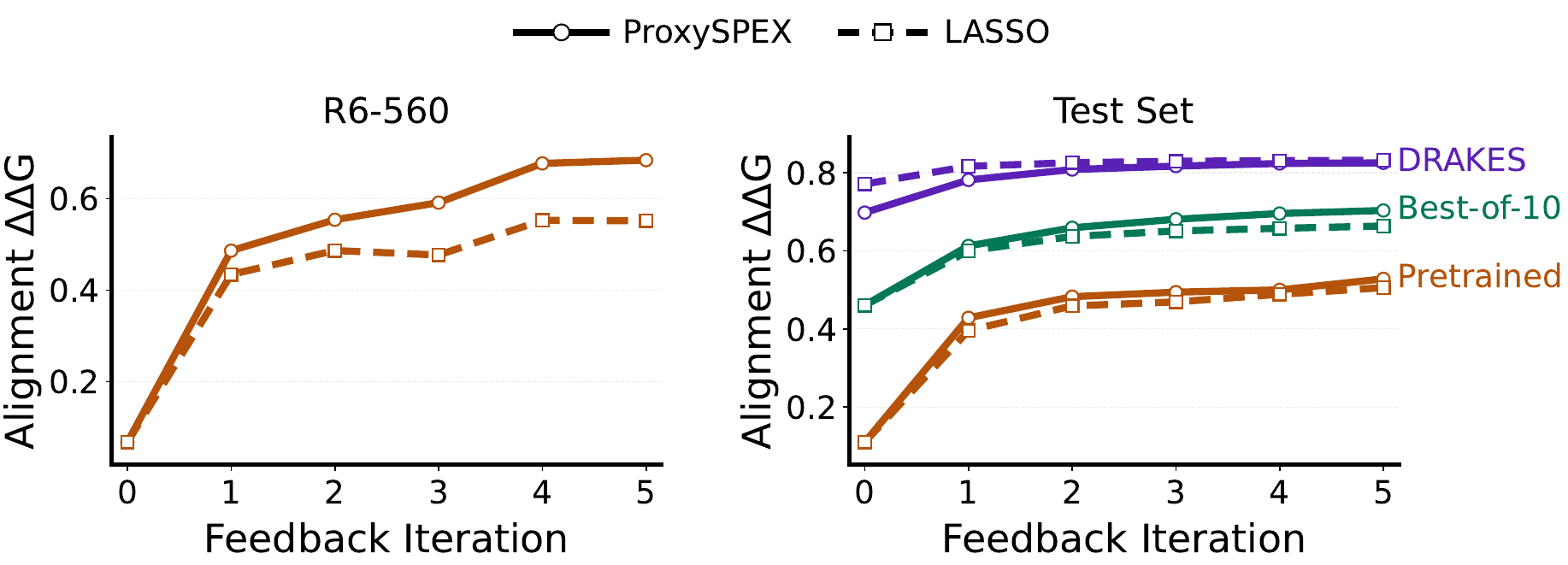}
    \caption{Spectral Feedback performance comparison with ProxySPEX vs first-order LASSO as the edit selection method. We constrain edit-selections to at most $k=20$ positions. \textbf{(left)} Visualization of one held-out feedback trajectory for the R6-560 protein backbone with the pre-trained model. \textbf{(right)} Average Spectral Feedback alignment across the test set, comparing ProxySPEX and LASSO as the edit-selection strategies for the pre-trained model, Best-of-10, and DRAKES.}
    \label{fig:placeholder3}
\end{figure}

\newpage
\subsection{Target Protein: R6-560 (r6\_560\_TrROS\_Hall)}
\label{app:protein_case_study_exp_r6_560}
\vspace{-1.5em}

\begin{center}
\[
\boxed{
\begin{aligned}
    &\text{Initial: SKPPKVVTVEVAVTKPDGKTELVKVTFTNLPRELKPGDTVTIPETGQKATVVKIIP}\\
\end{aligned}
}
\]
\end{center}

\begin{center}
\textbf{ProxySPEX} - $f_{avg}$
\end{center}
\vspace{-0.5em}

\textit{Iteration 1
}

Targets: \underline{4}, 6, \underline{9}, 12, \underline{15}, 18, 19, 20, 24, 25, \underline{26}, 28, 34, 40, \underline{41}, 43, 48, 49, \underline{50}, \underline{52}

Changes: 4 (K $\rightarrow$ R), 9 (E $\rightarrow$ V), 15 (P $\rightarrow$ A), 26 (F $\rightarrow$ L), 41 (I $\rightarrow$ L), 50 (V $\rightarrow$ I), 52 (K $\rightarrow$ E)

Result: {\small SKPPRVVTVVVAVTKADGKTELVKVTLTNLPRELKPGDTVTLPETGQKATIVEIIP}

\textit{Iteration 2
}

Targets: \underline{0}, \underline{1}, \underline{9}, 13, \underline{14}, 17, 18, 19, 25, 27, \underline{28}, 32, \underline{34}, 36, \underline{38}, 43, 46, \underline{47}, \underline{54}

Changes: 0 (S $\rightarrow$ A), 1 (K $\rightarrow$ P), 9 (V $\rightarrow$ L), 14 (K $\rightarrow$ R), 28 (N $\rightarrow$ G), 34 (K $\rightarrow$ R), 38 (T $\rightarrow$ V), 47 (K $\rightarrow$ E), 54 (I $\rightarrow$ L)

Result: {\small APPPRVVTVLVAVTRADGKTELVKVTLTGLPRELRPGDVVTLPETGQEATIVEILP}

\textit{Iteration 3}

Targets: 0, 8, 11, 12, 17, \underline{18}, 20, \underline{22}, \underline{23}, 28, 30, 36, 42, 43, 44, 45, \underline{47}, 49, 55

Changes: 18 (K $\rightarrow$ R), 22 (V $\rightarrow$ R), 23 (K $\rightarrow$ R), 47 (E $\rightarrow$ R)

Result: {\small APPPRVVTVLVAVTRADGRTELRRVTLTGLPRELRPGDVVTLPETGQRATIVEILP}

\textit{Iteration 4}

Targets: 0, \underline{7}, 8, 13, \underline{14}, 16, 19, \underline{20}, 28, 29, 30, 34, 35, 38, 40, 45, \underline{46}, \underline{47}, 48, 54

Changes: 7 (T $\rightarrow$ E), 14 (R $\rightarrow$ E), 20 (E $\rightarrow$ V), 46 (Q $\rightarrow$ E), 47 (R $\rightarrow$ E)

Result: {\small APPPRVVEVLVAVTEADGRTVLRRVTLTGLPRELRPGDVVTLPETGEEATIVEILP}

\textit{Iteration 5}

Targets: 0, 1, 6, 10, 16, 17, 24, 26, 29, 30, 31, 35, 36, 39, 42, \underline{47}, 51, 52, 54, 55

Changes: 47 (E $\rightarrow$ R)

Result: {\small APPPRVVEVLVAVTEADGRTVLRRVTLTGLPRELRPGDVVTLPETGERATIVEILP}

Reward Trajectory: [0.0679, 0.4864, 0.5541, 0.5915, 0.6773, 0.6843]

\begin{center}
\textbf{LASSO} - $f_{avg}$
\end{center}
\vspace{-0.5em}

\textit{Iteration 1}

Targets: \underline{0}, 1, \underline{4}, 6, 11, 13, 15, 18, 24, 25, \underline{26}, 27, 28, 29, 34, \underline{38}, \underline{41}, 48, \underline{50}, \underline{52}

Changes: 0 (S $\rightarrow$ A), 4 (K $\rightarrow$ R), 26 (F $\rightarrow$ L), 38 (T $\rightarrow$ V), 41 (I $\rightarrow$ L), 50 (V $\rightarrow$ I), 52 (K $\rightarrow$ R)

Result: {\small AKPPRVVTVEVAVTKPDGKTELVKVTLTNLPRELKPGDVVTLPETGQKATIVRIIP}

\textit{Iteration 2
}

Targets: \underline{1}, 2, 6, 10, 12, 13, \underline{15}, 18, \underline{23}, 24, \underline{27}, \underline{28}, 29, 34, 35, 36, 43, 44, 48

Changes: 1 (K $\rightarrow$ P), 15 (P $\rightarrow$ A), 23 (K $\rightarrow$ T), 27 (T $\rightarrow$ E), 28 (N $\rightarrow$ G)

Result: {\small APPPRVVTVEVAVTKADGKTELVTVTLEGLPRELKPGDVVTLPETGQKATIVRIIP}

\textit{Iteration 3}

Targets: 6, 12, \underline{14}, 17, 18, 24, \underline{27}, \underline{28}, \underline{29}, \underline{34}, 35, 36, 43, 44, 48, 52

Changes: 14 (K $\rightarrow$ R), 27 (E $\rightarrow$ R), 28 (G $\rightarrow$ D), 29 (L $\rightarrow$ T), 34 (K $\rightarrow$ R)

Result: {\small APPPRVVTVEVAVTRADGKTELVTVTLRDTPRELRPGDVVTLPETGQKATIVRIIP}

\textit{Iteration 4}

Targets: \underline{1}, 2, 5, 6, 8, 11, 12, 13, 18, 22, 24, \underline{27}, \underline{29}, \underline{34}, 36, 43, 44, 48, 52, 55

Changes: 1 (P $\rightarrow$ A), 27 (R $\rightarrow$ T), 29 (T $\rightarrow$ L), 34 (R $\rightarrow$ K)

Result: {\small AAPPRVVTVEVAVTRADGKTELVTVTLTDLPRELKPGDVVTLPETGQKATIVRIIP}

\textit{Iteration 5}

Targets: 3, 6, 10, 12, 13, 17, 24, 29, 30, \underline{34}, 36, 43, 44, 45, 48, 49

Changes: 34 (K $\rightarrow$ R)

Result: {\small AAPPRVVTVEVAVTRADGKTELVTVTLTDLPRELRPGDVVTLPETGQKATIVRIIP}

Reward Trajectory: [0.0679, 0.4345, 0.4868, 0.4770, 0.5528, 0.5518]

\subsection{Target Protein: 2KVV (v2K43S\_2KVV)}
\label{app:protein_case_study_exp_2kvv}
\vspace{-1.5em}
\begin{center}
\[
\boxed{
\begin{aligned}
    &\text{Initial: EKWIEQNELMKETGLKRSTITKLRKTKLKEGEHYKRVSKDGKPSKDATILYNLEKIKKLLK}\\
\end{aligned}
}
\]
\end{center}

\begin{center}
\textbf{ProxySPEX} - $f_{avg}$
\end{center}
\vspace{-0.5em}

\textit{Iteration 1
}

Targets: \underline{1}, \underline{5}, 6, 10, 19, 26, 28, 31, \underline{34}, 38, 41, 43, \underline{44}, 50, 53, 54, 56, \underline{57}, 59, 60

Changes: 1 (K $\rightarrow$ R), 5 (Q $\rightarrow$ E), 34 (K $\rightarrow$ R), 44 (K $\rightarrow$ P), 57 (K $\rightarrow$ E)

Result: {\small ERWIEENELMKETGLKRSTITKLRKTKLKEGEHYRRVSKDGKPSPDATILYNLEKIKELLK}

\textit{Iteration 2
}

Targets: 3, 6, 9, 10, 11, 15, 16, \underline{21}, 24, 26, 28, 31, 34, 41, \underline{42}, 48, 54, 55, \underline{56}, 60

Changes: 21 (K $\rightarrow$ R), 42 (P $\rightarrow$ D), 56 (K $\rightarrow$ L)

Result: {\small ERWIEENELMKETGLKRSTITRLRKTKLKEGEHYRRVSKDGKDSPDATILYNLEKILELLK}

\textit{Iteration 3}

Targets: 3, \underline{6}, 9, \underline{10}, \underline{11}, \underline{15}, 16, 17, \underline{24}, \underline{26}, \underline{28}, \underline{31}, 40, \underline{41}, 48, 51, 53, \underline{54}, 55, \underline{60}

Changes: 6 (N $\rightarrow$ R), 10 (K $\rightarrow$ A), 11 (E $\rightarrow$ A), 15 (K $\rightarrow$ A), 24 (K $\rightarrow$ R), 26 (K $\rightarrow$ R), 28 (K $\rightarrow$ E), 31 (E $\rightarrow$ R), 41 (K $\rightarrow$ R), 54 (K $\rightarrow$ A), 60 (K $\rightarrow$ A)

Result: {\small ERWIEERELMAATGLARSTITRLRRTRLEEGRHYRRVSKDGRDSPDATILYNLEAILELLA}

\textit{Iteration 4}

Targets: 1, 2, 8, 9, 14, \underline{15}, 19, \underline{20}, 21, 25, \underline{26}, 35, 36, \underline{38}, \underline{47}, 51, 52, 55, \underline{57}, 58

Changes: 15 (A $\rightarrow$ R), 20 (T $\rightarrow$ A), 26 (R $\rightarrow$ A), 38 (K $\rightarrow$ A), 47 (T $\rightarrow$ R), 57 (E $\rightarrow$ A)

Result: {\small ERWIEERELMAATGLRRSTIARLRRTALEEGRHYRRVSADGRDSPDARILYNLEAILALLA}

\textit{Iteration 5}

Targets: 0, 2, 8, 13, 14, \underline{15}, 16, \underline{19}, \underline{21}, 24, 25, 29, 36, 39, 43, 49, 52, 53, 58, 60

Changes: 15 (R $\rightarrow$ A), 19 (I $\rightarrow$ L), 21 (R $\rightarrow$ A)

Result: {\small ERWIEERELMAATGLARSTLAALRRTALEEGRHYRRVSADGRDSPDARILYNLEAILALLA}

Reward Trajectory: [-0.1766, 0.3299, 0.3596, 0.5910, 0.6261, 0.6717]

\begin{center}
\textbf{LASSO} - $f_{avg}$
\end{center}
\vspace{-0.5em}

\textit{Iteration 1}

Targets: \underline{5}, 6, 10, 12, 19, 25, 28, 29, 31, \underline{34}, 38, 40, 41, 44, 50, 54, 55, 56, \underline{57}, 60

Changes: 5 (Q $\rightarrow$ E), 34 (K $\rightarrow$ R), 57 (K $\rightarrow$ E)

Result: {\small EKWIEENELMKETGLKRSTITKLRKTKLKEGEHYRRVSKDGKPSKDATILYNLEKIKELLK}

\textit{Iteration 2
}

Targets: \underline{1}, 6, 9, 10, 15, 17, 19, 24, 26, 28, 30, \underline{31}, 34, \underline{38}, \underline{44}, 54, 55, 56, 57, 60

Changes: 1 (K $\rightarrow$ R), 31 (E $\rightarrow$ I), 38 (K $\rightarrow$ A), 44 (K $\rightarrow$ P)

Result: {\small ERWIEENELMKETGLKRSTITKLRKTKLKEGIHYRRVSADGKPSPDATILYNLEKIKELLK}

\textit{Iteration 3}

Targets: 3, \underline{6}, 9, 10, \underline{15}, 19, \underline{21}, \underline{24}, 26, 28, 29, 39, 43, 44, \underline{54}, 55, 56, \underline{60}

Changes: 6 (N $\rightarrow$ R), 15 (K $\rightarrow$ A), 21 (K $\rightarrow$ R), 24 (K $\rightarrow$ R), 54 (K $\rightarrow$ A), 60 (K $\rightarrow$ A)

Result: {\small ERWIEERELMKETGLARSTITRLRRTKLKEGIHYRRVSADGKPSPDATILYNLEAIKELLA}

\textit{Iteration 4}

Targets: 0, 3, 9, 10, 17, 19, 22, \underline{26}, 28, 29, \underline{42}, 44, 55, \underline{57}

Changes: 26 (K $\rightarrow$ R), 42 (P $\rightarrow$ D), 57 (E $\rightarrow$ A)

Result: {\small ERWIEERELMKETGLARSTITRLRRTRLKEGIHYRRVSADGKDSPDATILYNLEAIKALLA}

\textit{Iteration 5}

Targets: 0, 3, 9, \underline{10}, 17, 19, 28, 39, 42, 45, 55, 60

Changes: 10 (K $\rightarrow$ R)

Result: {\small ERWIEERELMRETGLARSTITRLRRTRLKEGIHYRRVSADGKDSPDATILYNLEAIKALLA}

Reward Trajectory: [-0.1766, 0.1106, 0.3755, 0.5710, 0.6129, 0.6203]



\end{document}